\documentclass[10pt,journal,compsoc]{IEEEtran}
\usepackage{amsmath,amsfonts}
\usepackage{algorithmic}
\usepackage{algorithm}
\usepackage{array}
\usepackage[caption=false,font=normalsize,labelfont=sf,textfont=sf]{subfig}
\usepackage{textcomp}
\usepackage{stfloats}
\usepackage{url}
\usepackage{verbatim}
\usepackage{graphicx}
\usepackage{cite}
\usepackage{enumitem}
\usepackage{tikz}
\usepackage{comment}
\usepackage{amsmath}
\usepackage{amsfonts}
\usepackage{bbding}
\usepackage{pgfplots}
\usepackage{booktabs}
\usepackage{diagbox}
\usepackage{makecell}
\usepackage{multirow}
\usepackage{xcolor}  
\usepackage{hhline}
\usepackage{colortbl}  

\newcommand{\eg}{\textit{e.g.}}
\usepackage{amssymb}
\usepackage{pifont}
\newcommand{\cmark}{\ding{51}}%
\newcommand{\xmark}{\ding{55}}%
\usepackage{dsfont}
\usepackage{tikz}
\usetikzlibrary{tikzmark}

\usepackage[hidelinks]{hyperref}
\hypersetup{
    colorlinks,
    linkcolor={red!100!black},
    citecolor={green!100!black},
    urlcolor={blue!100!black}
}

\begin{document}

\title{Semi-Supervised Unconstrained Head Pose Estimation in the Wild}

\author{Huayi Zhou, \IEEEmembership{Member, IEEE}, Fei Jiang, Jin Yuan, \\Yong Rui, \IEEEmembership{Fellow, IEEE}, Hongtao Lu, \IEEEmembership{Member, IEEE}, and Kui Jia, \IEEEmembership{Member, IEEE}
\thanks{H. Zhou and K. Jia (the corresponding author) are with School of Data Science, The Chinese University of Hong Kong, Shenzhen (e-mail: zhouhuayi@cuhk.edu.cn; kuijia@cuhk.edu.cn). F. Jiang is with Chongqing Academy of Science and Technology (e-mail: fjiang@mail.ecnu.edu.cn). J. Yuna and Y, Rui are with Lenovo Research and Technology (e-mail: yuanjin@seu.edu.cn; yongrui@lenovo.com). H. Lu is with Department of Computer Science and Engineering, Shanghai Jiao Tong University (e-mail: htlu@sjtu.edu.cn).}
\thanks{Manuscript received February 20, 2025; revised September 20, 2025.}}

\markboth{Journal of \LaTeX\ Class Files,~Vol.~14, No.~25, September~2025}%
{Shell \MakeLowercase{\textit{et al.}}: Semi-Supervised Head Pose Estimation}

\IEEEpubid{0000--0000/00\$00.00~\copyright~2025 IEEE}

\IEEEtitleabstractindextext{

\begin{abstract}
Existing research on unconstrained in-the-wild head pose estimation suffers from the flaws of its datasets, which consist of either numerous samples by non-realistic synthesis or constrained collection, or small-scale natural images yet with plausible manual annotations. This makes fully-supervised solutions compromised due to the reliance on generous labels. To alleviate it, we propose the first semi-supervised unconstrained head pose estimation method SemiUHPE, which can leverage abundant easily available unlabeled head images. Technically, we choose semi-supervised rotation regression and adapt it to the error-sensitive and label-scarce problem of unconstrained head pose. Our method is based on the observation that the aspect-ratio invariant cropping of wild heads is superior to previous landmark-based affine alignment given that landmarks of unconstrained human heads are usually unavailable, especially for underexplored non-frontal heads. Instead of using a pre-fixed threshold to filter out pseudo labeled heads, we propose dynamic entropy based filtering to adaptively remove unlabeled outliers as training progresses by updating the threshold in multiple stages. We then revisit the design of weak-strong augmentations and improve it by devising two novel head-oriented strong augmentations, termed pose-irrelevant cut-occlusion and pose-altering rotation consistency respectively. Extensive experiments and ablation studies show that SemiUHPE outperforms its counterparts greatly on public benchmarks under both the front-range and full-range settings. Furthermore, our proposed method is also beneficial for solving other closely related problems, including generic object rotation regression and 3D head reconstruction, demonstrating good versatility and extensibility. Code is in \url{https://github.com/hnuzhy/SemiUHPE}.
\end{abstract}
}
\maketitle

\section{Introduction}\label{intro}

\IEEEPARstart{H}{uman} head pose estimation (HPE) from a single RGB image in the wild is a long-standing yet still challenging problem \cite{murphy2008head, abate2022head}. It has numerous applications such as driver monitoring \cite{murphy2010head}, classroom observation \cite{ahuja2019edusense}, eye-gaze proxy \cite{ahuja2021classroom, nonaka2022dynamic}, and human intentions detection in social robots \cite{zapata2024guessing, singamaneni2024survey}. Meanwhile, HPE can also serve as a crucial auxiliary to facilitate other face or head related multi-tasks (\eg, landmark localization \cite{ranjan2017hyperface, wu2021synergy}, face alignment \cite{zhu2016face, guo2020towards, valle2020multi, albiero2021img2pose} and face shape reconstruction \cite{yu2018headfusion, martyniuk2022dad}).

\begin{figure}[!t]
	\centering
	\includegraphics[width=\linewidth]{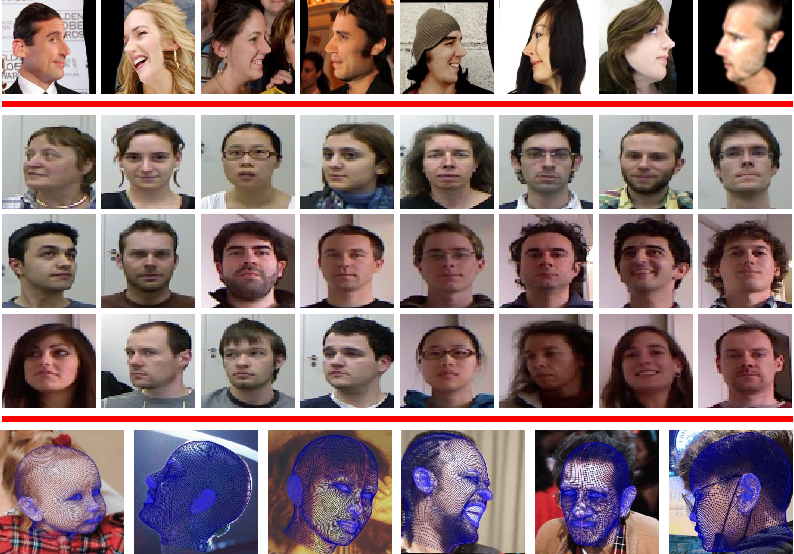}
	\vspace{-20pt}
	\caption{Examples of front-range datasets 300W-LP \cite{zhu2016face} (\textit{top}) having synthesized profile faces with many obvious artifacts and BIWI \cite{fanelli2013random} (\textit{middle}) collected in lab environments with only 24 sequences and very limited diversity, and full-range DAD-3DHeads \cite{martyniuk2022dad} (\textit{bottom}) with laboriously annotated 3D head mesh labels on 2D images.}
	\label{DatasetSamples}
	\vspace{-10pt}
\end{figure}

In the era of deep supervised learning, many HPE methods \cite{ruiz2018fine, yang2019fsa, zhou2020whenet, hempel20226d} need a large amount of labeled data to train. However, existing HPE datasets such as 300W-LP \cite{zhu2016face}, BIWI \cite{fanelli2013random} and DAD-3DHeads \cite{martyniuk2022dad} have their incompatible limitations for real applications. They are either artificially collected or synthesized \cite{zhu2016face, fanelli2013random, gu2017dynamic, kuhnke2019deep} so that having huge domain gaps and scarce diversities compared to natural heads. Others are manually annotated by certified experts \cite{martyniuk2022dad, wang20232dheadpose} at a significant time and economic cost with a small scale. Some examples are shown in Fig.~\ref{DatasetSamples}. In these labeled datasets, 300W-LP \cite{zhu2016face} covers yaw angles within $(-99^{\circ}, 99^{\circ})$ and BIWI \cite{fanelli2013random} within $(-75^{\circ}, 75^{\circ})$, both focusing mainly on front-range visible faces. In contrast, DAD-3DHeads \cite{martyniuk2022dad} officially reports 39\% front, 52\% side, and 9\% atypical poses, providing broader coverage but still limited representation of backward and invisible heads. These statistics confirm that unconstrained, full-range head pose estimation remains largely underexplored.

Affected by this, although many supervised learning methods designed and trained based on these datasets have achieved excellent quantitative performance on the test set (such as AFLW2000 \cite{zhu2016face} and BIWI \cite{fanelli2013random}), these models cannot be directly applied to the real world with complex scenarios. Some methods try to synthesize a large number of multi-view \cite{zeng20233d}, extreme-view \cite{dao2024efhq} or full-view \cite{an2023panohead} human heads to expand the training set, but these generated images still have hallucination defects and cannot be fully trusted. Different from most existing fully supervised methods, we turn to the semi-supervised learning (SSL) techniques \cite{tarvainen2017mean, berthelot2019mixmatch, sohn2020fixmatch, xie2020unsupervised}, and propose a semi-supervised unconstrained head pose estimation (SemiUHPE) method, that can leverage extensive easier obtainable yet unlabeled in-the-wild heads \cite{lin2014microsoft, yang2016wider, shao2018crowdhuman, kuznetsova2020open} in addition to partially labeled data to improve performance and generalization. The overall framework of our method is illustrated in Fig.~\ref{Framework}. It can not only avoid the laborious annotation of 3D head pose on 2D images which itself is ill-posed, but also greatly promote the estimation accuracy of challenging cases in real environments. 

\begin{figure*}[!t]
	\centering
	\includegraphics[width=\linewidth]{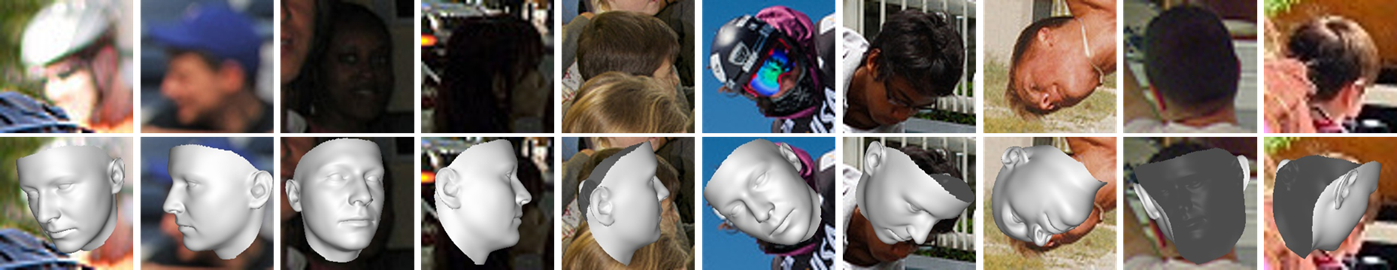}
	\vspace{-20pt}
	\caption{Our unconstrained head pose estimation results on wild challenging heads (\eg, heavy blur, extreme illumination, severe occlusion, atypical pose, and invisible face). Images are all selected from the COCO \cite{lin2014microsoft} dataset without head pose labels.}
	\vspace{-10pt}
	\label{Teaser}
\end{figure*}

Broadly speaking, our work is inspired by the recent proposed semi-supervised rotation regression method FisherMatch \cite{yin2022fishermatch, yin2025towards} focusing on generic objects. FisherMatch inherits two widely used paradigms in semi-supervised classification: the Mean-Teacher \cite{tarvainen2017mean} framework and pseudo label filtering devised by FixMatch \cite{sohn2020fixmatch}. Its main contribution is to use the entropy of prediction with the matrix Fisher distribution \cite{levinson2020analysis, mohlin2020probabilistic} as a measure for filtering pseudo labels, which resembles the classification confidence and enables it to handle semi-supervised rotation regression similar to FixMatch. Following this, we focus on the head pose estimation, which is a typical case of the general object rotation regression task. It should be emphasized that this is a quite non-trivial problem. Because comparing to general objects, the HPE task often requires a more sophisticated understanding of 3D spatial relationships and precise numerical output of yaw, pitch, and roll Euler angles. Besides, many less-explored challenging frontal faces and never-touched backwards heads cannot be well-solved by FisherMatch or any other alternative so far. Please refer Fig.~\ref{Teaser}, Fig.~\ref{Comparing2} and Fig.~\ref{ComparingDAD} for a cursory glance.

To this end, we aim to address unconstrained head pose estimation (UHPE) of omnidirectionality, including heads with common front-range angles and face-invisible back-range angles. We believe that only by solving the problem at any head orientation can we build fundamental HPE algorithms to support the prosperity and progress of downstream applications. Specifically, to tackle the semi-supervised UHPE task, we mainly propose the following three strategies for further improvements:

\textbf{(1) Aspect-Ratio Invariant Cropping.} We observe that previous methods \cite{ruiz2018fine, yang2019fsa, zhou2020whenet, hempel20226d} require aligned faces as inputs, which relies on pre-annotated landmarks. However, this does not apply to our task as back-range heads cannot be aligned and unlabeled data has no landmark labels. Moreover, face alignment during training can cause the affine deformation, thereby hindering the inference on natural heads. Therefore, we recommend using the landmark-free pre-processing of head cropping to maintain aspect-ratio invariant and enhance practical generalizability.

\textbf{(2) Dynamic Entropy-based Filtering.}  A key design of FixMatch is the confidence-based pseudo label filtering. FisherMatch develops the prediction entropy-based version for rotation regression. Although a pre-fixed threshold is validated to be effective, dynamic thresholds often make more sense in SSL. Such as adoption of curriculum pseudo labeling \cite{zhang2021flexmatch}, label grouping \cite{nassar2021all} and adaptive threshold \cite{wu2023chmatch}. For our task, due to the intermixing of hard and noisy samples in unlabeled wild heads, we consider that gradually updating the threshold as the training converges is a better and more general choice.

\textbf{(3) Head-Oriented Strong Augmentations.} Another key idea of FixMatch is the weak-strong paired augmentations, which feeds the teacher model by weakly augmented unlabeled inputs to guide the student model fed by the same unlabeled yet strongly augmented inputs. Based on it, some SSL methods for detection \cite{kim2022mum}, segmentation \cite{yang2023revisiting} and keypoints \cite{xie2021empirical, huang2023semi} found that an advanced strong augmentation is quite important. Similarly, we re-examine the properties of unlabeled heads and invent two novel superior augmentations: pose-irrelevant cut-occlusion and pose-altering rotation consistency.

The three above-mentioned strategies are complementary: cropping guarantees a stable and scale-invariant input, dynamic filtering ensures the reliability of semi-supervised training, and augmentations promote robustness against extreme pose and occlusion variations. Their synergy enables our framework to consistently improve accuracy and generalization in unconstrained HPE.
In method section, we provide more details of them. We also present how these strategies can be utilized to solve the task generic object rotation regression similar to HPE and help improve another task 3D head reconstruction strongly related to HPE. Despite the simplicity, these seamless adaptations reveal valuable universality and scalability of SemiUHPE. In experiments, we demonstrate the effectiveness and versatility of our proposed strategies by extensive comparison and ablation studies. In addition, we give impressive qualitative estimation results on challenging wild heads, which is promising for downstream applications.

To sum up, we mainly have five contributions: (1) The semi-supervised unrestricted head pose estimation (SemiUHPE) task for wild RGB images is proposed for the first time. (2) A well-performed framework including novel customized strategies for tackling the SemiUHPE problem is proposed. (3) Benchmarks with many baseline methods for the SemiUHPE are constructed. (4) New SOTA results are achieved under both front-range and full-range HPE settings. (5) Our proposed strategies are verified to be effortlessly applicable to promote two other tasks similar or relevant to unconstrained HPE.

Our content is organized as follows: Firstly, in Section~\ref{relwork}, we present related works including various 2D head pose estimation methods, the popular semi-supervised learning algorithms, and some semi-supervised rotation regression researches about general objects. Then, in Section~\ref{method}, we give a definition of the problem, the description of overall SemiUHPE framework, motivations, explanations of proposed strategies and adaptations for other tasks. After that, in Section~\ref{exps}, we show multiple datasets and settings required for training, corresponding implementation details, quantitative result comparisons under various settings, complete ablation studies, qualitative visualization results and other optional setup attempts. Finally, in Section~\ref{conclusion}, we summarize findings of this paper and provide an in-depth discussion of some future directions.

\section{Related Work}\label{relwork}

\subsection{RGB-based Head Pose Estimation}
Human head pose estimation (HPE) using monocular RGB images is a widely researched field \cite{murphy2008head, abate2022head}. Benefiting from deep CNN, data-driven supervised learning methods tend to dominate this field. Basically, we can divide them into four categories based on landmarks \cite{kazemi2014one, bulat2017far, zhu2016face, deng2020retinaface, cantarini2022hhp}, Euler angles \cite{ruiz2018fine, yang2019fsa, zhou2020whenet, zhang2020fdn, geng2020head, wang20232dheadpose, zhou2023directmhp}, vectors \cite{hsu2018quatnet, cao2021vector, liu2021mfdnet, dai2020rankpose, albiero2021img2pose, hempel20226d, zhang2023tokenhpe} or 3D Morphable Model (3DMM) \cite{zhu2016face, guo2020towards, ruan2021sadrnet, wu2021synergy, martyniuk2022dad, kao2023towards, li2023dsfnet}. Euler angles-based methods are essentially hindered by the \textit{gimbal lock}. Vectors-based methods using representations such as unit quaternion \cite{hsu2018quatnet}, rotation vector \cite{albiero2021img2pose} and rotation matrix \cite{liu2021mfdnet, hempel20226d, zhang2023tokenhpe} can alleviate this drawback and allow full-range predictions. The 3DMM-based methods treat HPE as a sub-task with optimizing multi-tasks when doing 3D face reconstruction, which currently keep SOTA performance on both front-range \cite{wu2021synergy, li2023dsfnet} and full-range \cite{martyniuk2022dad} HPE. In this paper, we focus on full-range unconstrained HPE and choose rotation matrix as the pose representation.

\subsection{Semi-Supervised Learning (SSL)}
The SSL aims to improve models by exploiting a small-scale labeled data and a large-scale unlabeled data. It can be categorized into pseudo-label (PL) based \cite{radosavovic2018data, oliver2018realistic, sohn2020fixmatch, nassar2021all, wu2023chmatch, nassar2023protocon} and consistency-based \cite{laine2016temporal, tarvainen2017mean, berthelot2019mixmatch, xie2020unsupervised, zhang2021flexmatch, hu2021simple}. The PL-based method selects unlabeled images into the training data iteratively by utilizing suitable thresholds to filter out uncertain samples with low-confidence. While, the consistency-based method enforces outputs or intermediate features to be consistent when the input is randomly perturbed. For example, MixMatch \cite{berthelot2019mixmatch} combines consistency regularization with entropy minimization to obtain confident predictions. Based on MixMatch, SimPLE \cite{hu2021simple} exploits similar high confidence pseudo labels. FixMatch \cite{sohn2020fixmatch} integrates both pseudo label filtering and weak-to-strong augmentation consistency. FlexMatch \cite{zhang2021flexmatch}, CCSSL \cite{yang2022class} and FullMatch \cite{chen2023boosting} extend FixMatch by adopting the curriculum pseudo labeling, contrastive learning and usage of all unlabeled data, respectively. We follow the empirically powerful FixMatch family \cite{zhang2021flexmatch, yang2022class, chen2023boosting, wu2023chmatch} to propose novel solutions for tackling HPE.

\subsection{Semi-Supervised Rotation Regression}
This is a less-studied field compared with other popular SSL tasks such as classification and detection. For the 6D pose estimation, Self6D \cite{wang2020self6d, wang2021occlusion} establishes self-supervision for 6D pose by enforcing visual and geometric consistencies on top of unlabeled RGB-D images. NVSM \cite{wang2021neural} builds a category-level 3D cuboid mesh for estimating pose of rigid object in a synthesis-and-matching way. Recently, based on FixMatch \cite{sohn2020fixmatch}, FisherMatch \cite{yin2022fishermatch} firstly proposes the semi-supervised rotation regression for generic objects. It can automatically learn uncertainties along with predictions by introducing the matrix Fisher distribution \cite{levinson2020analysis, mohlin2020probabilistic} to build a probabilistic model of rotation. The entropy of this distribution has been validated to be an efficient indicator of prediction for pseudo label filtering. The improved version \cite{yin2025towards} has integrated the rotation Laplace distribution \cite{yin2022laplace} which is more robust to the disturbance of outliers and enforces much gradient to the low-error region. Essentially, HPE is subordinate to the rotation regression problem. MFDNet \cite{liu2021mfdnet} also utilizes the matrix Fisher distribution to model head rotation uncertainty. Kuhnke et al. \cite{kuhnke2021relative, kuhnke2023domain} introduce the relative pose consistency into semi-supervised constrained HPE on dataset BIWI, while we explicitly target the more challenging unconstrained in-the-wild setting with novel cropping, dynamic filtering, and head-oriented augmentations. We follow \cite{yin2022fishermatch, liu2021mfdnet} and adopt this distribution to tackle the SemiUHPE task.

\begin{figure*}
	\centering
	\includegraphics[width=\textwidth]{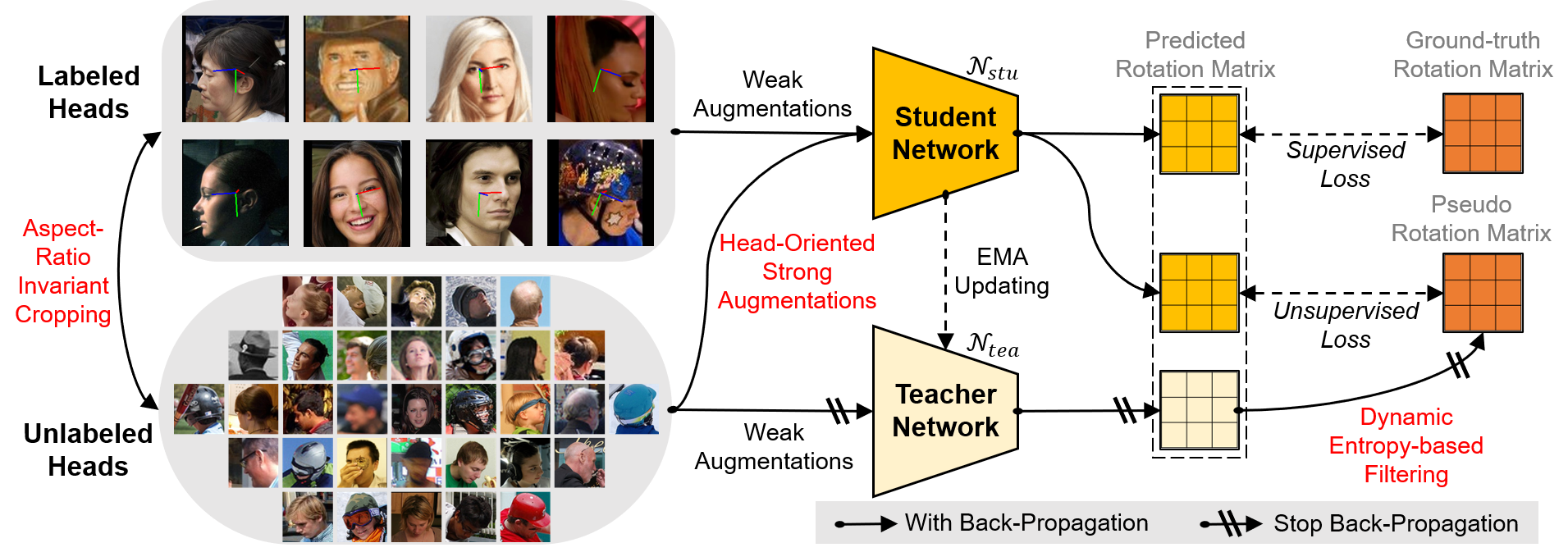}
	\vspace{-20pt}
	\caption{The framework illustration of our SemiUHPE. We leverage small-scale labeled heads and large-scale unlabeled wild heads to optimize the teacher-student mutual learning Mean-Teacher framework. Three customized strategies are marked with red color. We finally keep the student model which is more efficient and robust for HPE evaluation.}
	\label{Framework}
	\vspace{-10pt}
\end{figure*}

\section{Method}\label{method}

\textbf{Problem Definition.}
Our SemiUHPE aims to utilize a small set of head images with pose labels $\mathcal{D}^l\!=\!\{(\mathbf{x}^l_i, \mathbf{y}^l_i)\}_{i=1}^{N_l}$ and fully explore a large set of unlabeled head images $\mathcal{D}^u\!=\!\{\mathbf{x}^u_i\}_{i=1}^{N_u}$. Here, $\mathbf{x}^l$ and $\mathbf{x}^u$ represent labeled and unlabeled RGB images respectively, and $\mathbf{y}^l_i$ is the ground-truth head pose label of $\mathbf{x}^l$ such as a rotation matrix or three Euler angles. The $N_l$ and $N_u$ are the number of labeled and unlabeled head images, respectively. Usually, the labeled set $\mathcal{D}^l$ contains either many accurate annotations yet non-photorealistic images (\eg, 300W-LP \cite{zhu2016face}), or laboriously hand-annotated labels but limited images (\eg, DAD-3DHeads \cite{martyniuk2022dad}). The unlabeled set $\mathcal{D}^u$ has much more realistic wild heads with spontaneous expressions and diversified properties, such as COCO \cite{lin2014microsoft}. In short, for those challenging heads in the unlabeled set which is defined as $\mathcal{D}^u\!=\!\{ \mathcal{D}^u_{id} \bigcup \mathcal{D}^u_{ood} \}$, we need to exploit the in-distribution valuable portion $\mathcal{D}^u_{id}$ that is less-explored, and avoid negative impact of the noisy out-of-distribution portion $\mathcal{D}^u_{ood}$.

\textbf{Clarification of $\mathcal{D}^u_{id}$ and $\mathcal{D}^u_{ood}$.} In this work, the terms $id$ and $ood$ are not intended to strictly follow the classical notions of covariate or distributional shift, but rather serve as task-specific operational definitions. Concretely, $\mathcal{D}^u_{id}$ denotes samples that, despite possibly having large head rotations, still provide sufficient discriminative cues (\eg,, side-view heads with visible contours or head-top structure) such that the network prediction confidence remains relatively high. These heads are valuable for improving robustness although they are rarely covered by existing labeled datasets. By contrast, $\mathcal{D}^u_{ood}$ corresponds to extremely ambiguous cases with multiple sources of visual degradation (\eg, backward heads with heavy occlusion or blur), where even humans may struggle to infer orientation. Such cases are regarded as noisy and are progressively filtered by our dynamic entropy-based strategy.

\subsection{Framework Overview}

As shown in Fig.~\ref{Framework}, we adopt the Mean-Teacher \cite{tarvainen2017mean} in our overall framework. The teacher model $\mathcal{N}_{tea}$ is the exponential moving average (EMA) of the student model $\mathcal{N}_{stu}$. While, the $\mathcal{N}_{stu}$ is supervisedly trained by labeled data, and also unsupervisedly guided by pseudo labels of unlabeled data predicted by the $\mathcal{N}_{tea}$. In this way, two models are enforced by the history consistency. Then, FixMatch \cite{sohn2020fixmatch} extends it by combining weak-strong augmentations and pseudo label filtering for classification. For rotation regression, FisherMatch \cite{yin2022fishermatch} follows FixMatch and utilizes the \textit{entropy} of predicted \textit{matrix Fisher distribution} as a measure for pseudo label filtering. We briefly review this probabilistic rotation distribution below.

\textbf{Matrix Fisher Distribution $\mathcal{MF}(\mathbf{R};\mathbf{A})$}
This is a favorable probabilistic modeling of deep rotation estimation, which has a bounded gradient \cite{mohlin2020probabilistic, liu2021mfdnet} and intrinsic advantage than its counterpart Bingham distribution \cite{levinson2020analysis}. Specifically, it is a distribution over $\mathcal{SO}$(3) with probability density function as:
\begin{equation}\small
	p(\mathbf{R}) = \mathcal{MF}(\mathbf{R};\mathbf{A}) = \frac{1}{F(\mathbf{A})}\exp(\text{tr}(\mathbf{A}^T\mathbf{R}))
\end{equation}
where $\mathbf{A}\!\in\!\mathbb{R}^{3\times3}$ is an arbitrary $3\!\times\!3$ matrix and $F(\mathbf{A})$ is the normalizing constant. Then, the mode $\mathbf{R}$ and dispersion $\mathbf{S}$ of the distribution are computed as:
\begin{equation}\small
	\mathbf{R} = \mathbf{U} \left[ \begin{array}{ccc} 1&\quad0&\quad0 \\ 0&\quad1&\quad0 \\ 0&\quad0&\quad\det(\mathbf{U}\mathbf{V}) \end{array} \right] \mathbf{V}^T
	\label{ModeDispersion}
\end{equation}
where $\mathbf{U}$ and $\mathbf{V}$ are from the singular value decomposition (SVD) of $\mathbf{A}\!=\!\mathbf{U} \mathbf{S} \mathbf{V}^T$. Each singular value $s_i$ in $\mathbf{S}\!=\!\text{diag}(s_1, s_2, s_3)$ is sorted in descending order, and indicates the concentration strength. 

\textbf{Entropy-based Confidence Measure}
During training, the network regressor $\mathcal{N}$ takes a single RGB image $\mathbf{x}$ as input and outputs a $3\!\times\!3$ matrix $\mathbf{A}_f\!=\!\mathcal{N}(\mathbf{x})$. The predicted $\mathbf{A}_f$ is a matrix Fisher distribution $f\!\sim\!\mathcal{MF}(\mathbf{A}_f)$. It contains a predicted rotation and the information of concentration by computing mode $\mathbf{R}_f$ and dispersion $\mathbf{S}_f$ as in Eq.~\ref{ModeDispersion}, respectively. The \textit{entropy} of this prediction, used as confidence measure of uncertainty, can be computed as:
\begin{equation}\small
	H(f) = \log{F_f} - \sum\nolimits^4_{i=1}\left({z_{f_i} \frac{1}{F_f} \frac{\partial{F_f}}{\partial{z_{f_i}}}}\right)
	\label{EntropyCal}
\end{equation}
where $F_f$ is constant wrt. parameter $\mathbf{Z}\!=\!\text{diag}(0, z_1, z_2, z_3)$. And $\mathbf{Z}$ is a $4\!\times\!4$ diagonal matrix with $0\!\geq\!z_1\!\geq\!z_2\!\geq\!z_3$. The element $z_i$ is from the corresponding unit quaternion $\mathbf{q}\!\in\!\mathcal{S}^3$. Assume $\mathbf{A}_f\!=\!\mathbf{U}_f \mathbf{S}_f \mathbf{V}_f^T$, $\gamma$ is the standard transform from unit quaternion to rotation matrix, $\mathbf{e}_i$ is the $i$-th column of $\mathbf{I}_4$, and $\mathbf{E}_i\!=\!\gamma(\mathbf{e}_i)$, then $z_{f_i}$ is the trace of $\mathbf{E}_i^T \mathbf{S}_i$. More details of the derivation are in \cite{mohlin2020probabilistic, yin2022fishermatch}. Generally, a lower entropy indicates a more peaked distribution that means less uncertainty and higher confidence. Next, we discuss how to optimize the primary entropy-based filtering strategy for SemiUHPE, as well as design stronger augmentations for unlabeled images $\mathbf{x}^u$.

\subsection{Aspect-Ratio Invariant Cropping}
\begin{figure}
	\centering
	\includegraphics[width=\columnwidth]{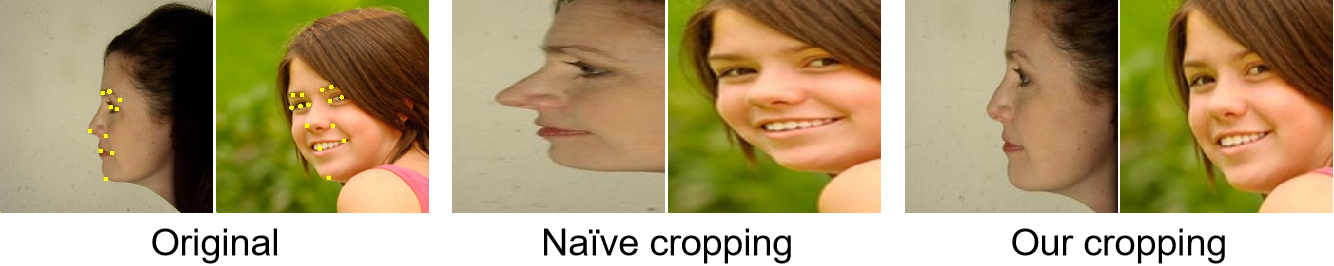}
	\vspace{-20pt}
	\caption{The illustration of how a naive cropping-resizing leads to perceived orientation and affine deformation.}
	\label{HeadCropping}
\end{figure}

For input preprocessing, we call for keeping the aspect-ratio invariant by loosely cropping head-centered images with bounding boxes and padding the out-of-plane area with zero. The reasons are two-fold. Firstly, the naive cropping-resizing may lead to scaling-related flaws \cite{mohlin2020probabilistic} such as perceived orientation change. Secondly, existing HPE methods \cite{ruiz2018fine, yang2019fsa, zhou2020whenet, hempel20226d} take it for grant that aligning face with landmarks will bring better results like face recognition \cite{liu2017sphereface, deng2019arcface}. However, this often introduces severe affine deformation that disrupts the natural face (refer Fig.~\ref{HeadCropping}). Moreover, face landmarks are not applicable to back-range or wild-collected heads. We will verify the necessity and advantages of aspect-ratio invariant cropping in experiments.

\begin{figure}
	\centering
	\includegraphics[width=\columnwidth]{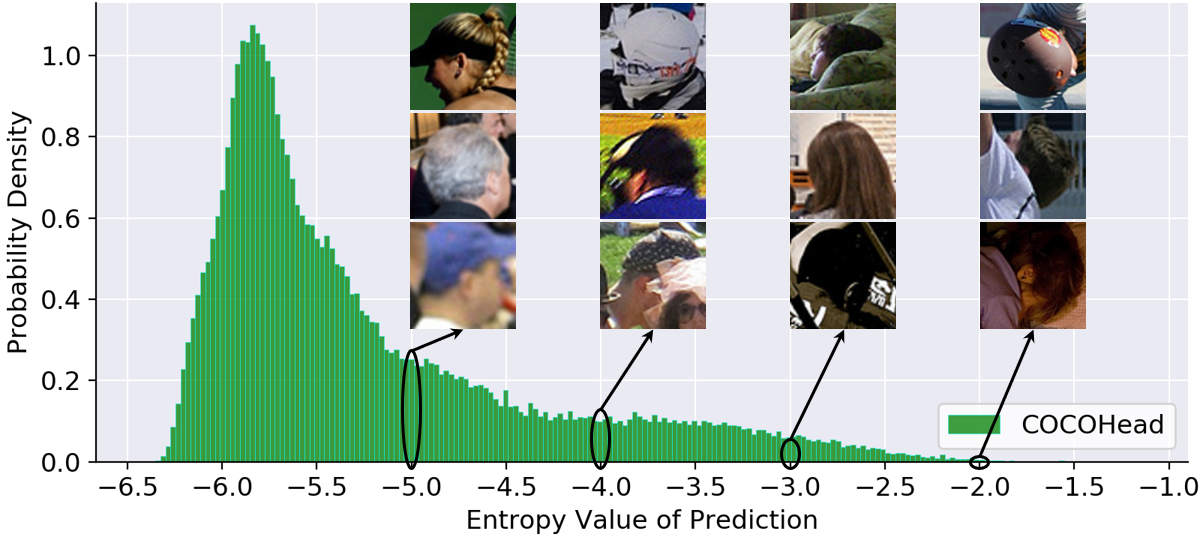}
	\vspace{-20pt}
	\caption{The illustration of prediction entropies corresponding to their head samples in the unlabeled dataset (\eg~COCOHead).}
	\label{EntropStat}
	\vspace{-10pt}
\end{figure}

\subsection{Dynamic Entropy-based Filtering}
In original FisherMatch, it adopts a trivial entropy-based filtering rule on unlabeled data which keeps the prediction as a pseudo label when its entropy is lower than a pre-fixed threshold $\tau$. The obtained unsupervised loss is:
\begin{equation}\small
	\mathcal{L}_{unsup}(\mathbf{x}^u) = \mathds{1}_{(H(p_{tea}) \leq \tau)} \mathcal{L}^{CE}(p_{tea}, p_{stu})
	\label{UnsupLoss}
\end{equation}
where $\mathds{1}_{(\cdot)}$ is the indicator function, being 1 if the condition holds and 0 otherwise. $H(p_{tea})$ is the prediction entropy computed as in Eq.~\ref{EntropyCal}. $\mathcal{L}^{CE}(\cdot, \cdot)$ is the cross entropy loss to enforce consistency between two continuous matrix Fisher distributions \cite{mohlin2020probabilistic}. The $p_{tea}$ and $p_{stu}$ are denoted as $p_{tea}\!=\!\mathcal{MF}(\mathbf{A}^u_{tea})$ and $p_{stu}\!=\!\mathcal{MF}(\mathbf{A}^u_{stu})$ which are outputted by the teacher model $\mathbf{A}^u_{tea}\!=\!\mathcal{N}_{tea}(\mathbf{x}^u)$ and the student model $\mathbf{A}^u_{stu}\!=\!\mathcal{N}_{stu}(\mathbf{x}^u)$, respectively.

For our task, the unlabeled set $\mathcal{D}^u$ has many challenging heads. It is quite difficult to distinguish whether a sample in $\mathcal{D}^u$ belongs to $\mathcal{D}^u_{id}$ or $\mathcal{D}^u_{ood}$ for the teacher model through a fixed threshold. For example, as shown in Fig.~\ref{EntropStat}, we calculated the predicted entropies of the teacher model $\mathcal{N}_{tea}$ for all samples in $\mathcal{D}^u$. The $\mathcal{N}_{tea}$ has quite certain predictions for most samples (lower entropies). While, samples with high uncertainty (higher entropies) are divided into two types: hard heads still in $\mathcal{D}^u_{id}$ or noisy heads in $\mathcal{D}^u_{ood}$. The former includes cases with severe occlusion or atypical pose which are infrequent in $\mathcal{D}^l$ yet possible to be correctly predicted. The latter contains extreme noisy cases with unrecognizable pose due to insufficient context or incorrect category. Moreover, the predictive ability of $\mathcal{N}_{tea}$ improves with the deepening of training, which means the difficulty and uncertainty of the same sample for $\mathcal{N}_{tea}$ is also changing.

Therefore, we propose the dynamic entropy-based filtering to improve the pseudo-label quality and enhance the model's robustness in real-world. Specifically, with the assumption that $\mathcal{D}^u\!=\!\{ \mathcal{D}^u_{id} \bigcup \mathcal{D}^u_{ood} \}$, we choose to retain a portion of unlabeled data in $\mathcal{D}^u$ for unsupervised training. For each mini-batch inputs, we still need a concrete threshold $\tau_k$ to filter predictions, where $\tau_k$ is progressively updated throughout $K$ stages. The $\tau_t$ is calculated as:
\begin{equation}\small
	\tau_k = \mathsf{percentile} \langle H(\mathcal{MF}(\mathcal{N}_{tea}^k(\mathbf{x}^u_i)))|_{i=1}^{N_u}, \delta \rangle
\end{equation}
where $\delta$ is the percentage of remained unlabeled data, and closely related to the unknown $\mathcal{D}^u_{ood}$ in $\mathcal{D}^u$. $\mathsf{percentile\langle \cdot,\cdot \rangle}$ returns the value of $\delta^{th}$ percentile. $\mathcal{N}_{tea}^k$ is the teacher model in $k$-th stage ($k \in \{1,2,...,K\}$). Then, we revise Eq.~\ref{UnsupLoss} as:
\begin{equation}\small
	\mathcal{L'}_{unsup}(\mathbf{x}^u) = \mathds{1}_{(H(p^k_{tea}) \leq \tau_k)} \mathcal{L}^{CE}(p^k_{tea}, p_{stu})
	\label{UnsupLossNew}
\end{equation}
where $p^k_{tea}\!=\!\mathcal{MF}(\mathcal{N}_{tea}^k(\mathbf{x}^u))$. Usually, with a given percentage $\delta$, the dynamic entropy threshold $\tau_k$ will decrease as the stage $k$ increases. And the optimal value of $\delta$ is inversely proportional to the images number of $\mathcal{D}^u_{ood}$ in $\mathcal{D}^u$. It should be emphasized that in our context, the distinction between $\mathcal{D}^u_{id}$ and $\mathcal{D}^u_{ood}$ reflects the difficulty and reliability of pose inference rather than classical covariate shift, which helps the dynamic threshold adaptively retain plausible but hard samples while filtering extremely noisy ones. Representative $\mathcal{D}^u_{id}$ and $\mathcal{D}^u_{ood}$ examples under our definition are visualized in Figs.~\ref{EntropStat}, \ref{Comparing1}, \ref{Comparing2}, \ref{FailureCases} to illustrate the filtering effect in practice.

\subsection{Head-Oriented Strong Augmentations}
FisherMatch follows the original weak-strong augmentations in FixMatch, and defines both the weak augmentation $T_\mathsf{weak}$ and strong augmentation $T_\mathsf{strong}$ as random cropping-resizing only with different scale factors. Empirically, many SSL methods \cite{xie2020unsupervised, xie2021empirical, huang2023semi} have found that the core of weak-strong augmentations paradigm in FixMatch is a more advanced strong augmentation than $T_\mathsf{strong}$. We thus propose two novel strong augmentations for unlabeled heads.

\begin{figure}[t]
	\centering
	\includegraphics[width=\columnwidth]{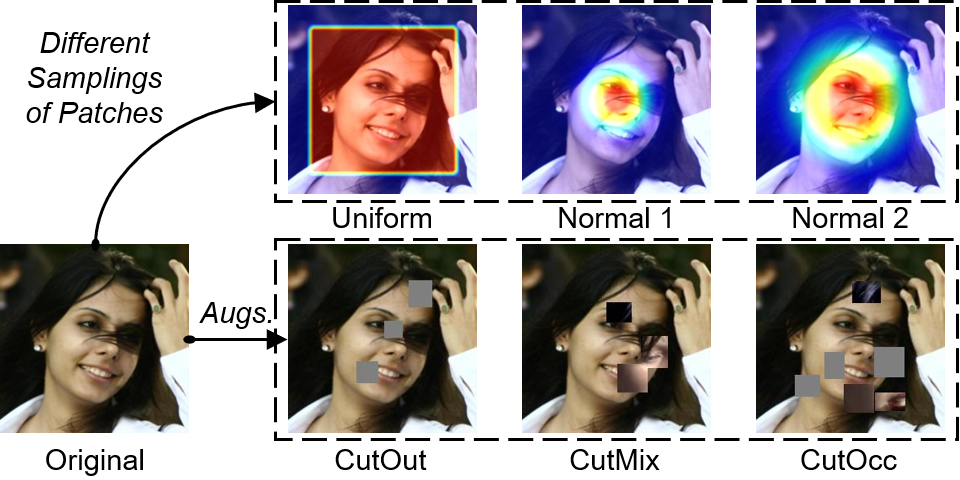}
	\vspace{-20pt}
	\caption{The illustration of patches sampling with different distributions, and the novel pose-irrelevant cut-occlusion (CutOcc) augmentation.}
	\label{AugsCOCM}
	\vspace{-10pt}
\end{figure}

\begin{figure}[t]
	\centering
	\includegraphics[width=\columnwidth]{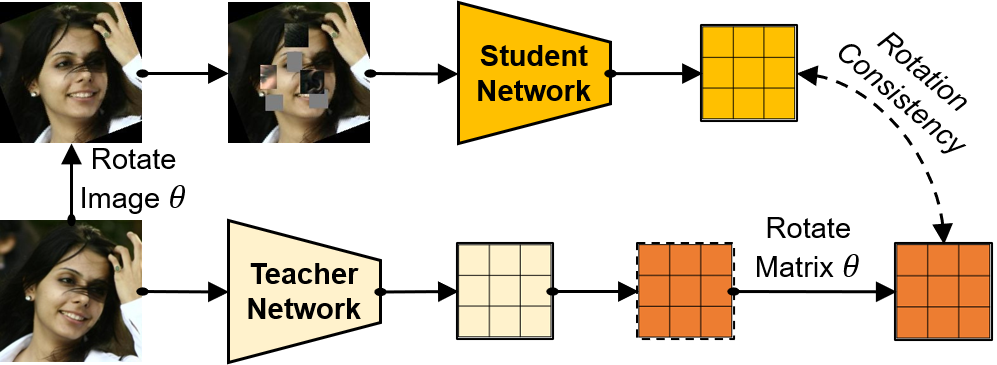}
	\vspace{-20pt}
	\caption{The illustration of proposed pose-altering rotation consistency (RotCons) augmentation. CutMix-related patches are fetched from other samples in the same mini-batch.}
	\label{AugsRotCons}
	\vspace{-10pt}
\end{figure}

\subsubsection{Pose-Irrelevant Cut-Occlusion}
Firstly, we explore pose-irrelevant augmentations such as popular CutOut \cite{devries2017improved}, Mixup \cite{zhang2018mixup} and CutMix \cite{yun2019cutmix}. CutOut simulates random occlusion. Mixup combines global features in different samples. CutMix balances both occlusion and crossed features. Considering that self- or emerged-occlusion is common in wild heads, we extend the CutOut and CutMix by sampling target patches with head-centered distributions. Heuristically, we provide three proposals of sampling distribution $\mathcal{S}$: Uniform distribution with a small distance from the boundary ($\mathcal{S}_\mathsf{Uniform}$), Normal distribution with a smaller variance ($\mathcal{S}_\mathsf{Normal1}$) and a larger variance ($\mathcal{S}_\mathsf{Normal2}$). Among them, we verified in experiments that the $\mathcal{S}_\mathsf{Normal2}$ is superior for its stronger head-centered concentration of noise addition. Then, we propose to conduct CutOut and CutMix in sequence to obtain an advanced combination named Cut-Occlusion (CutOcc). The motivation is to leverage the synergistic effect between two existing components. As shown in Fig.~\ref{AugsCOCM}, CutOcc is also visually understandable.

\subsubsection{Pose-Altering Rotation Consistency}
Although HPE is sensitive to rotation, we can still perform in-plane rotation augmentation in $\mathcal{SO}$(3) similar to SSL keypoints detection \cite{rhodin2018unsupervised, xie2021empirical, huang2023semi}. For a batch of unlabeled heads $\{\mathbf{x}^u_i\}^{B_u}_{i=1}$, we present unsupervised rotation consistency augmentation as shown in Fig.~\ref{AugsRotCons}. On one hand, we rotate each $\mathbf{x}^u_i$ with a random angle $\theta$ from $(-30^\circ, 30^\circ)$, with selectively conducting the subsequent pose-irrelevant operation CutOcc. The strongly augmented $\overline{\mathbf{x}^u}$ are then fed into the student model $\mathcal{N}_{stu}$, which outputs parameters in the form of matrix $\overline{p^i_{stu} }\in\mathbb{R}^{3\times3}$. On the other hand, we directly feed weakly augmented $\widetilde{\mathbf{x}^u}$ into the teacher model $\mathcal{N}_{tea}$, and obtain the predicted matrix $\widetilde{p^i_{tea}}$. Then, we need to rotate $\widetilde{p^i_{tea}}$ around the Z-axis with corresponding degree for consistency training. We summarize these steps as follows.
\begin{equation}\small
\begin{aligned}
	& \overline{p^i_{stu}} = \mathcal{MF}(\mathcal{N}_{stu}(\overline{\mathbf{x}^u_i})), \quad
	\overline{\mathbf{x}^u_i} = T_\mathsf{CutOcc}(T_{\mathsf{Rot}_\theta}(\mathbf{x}^u_i)) \\
	& \widetilde{p^i_{tea}} = \mathcal{MF}(\mathcal{N}_{tea}(\widetilde{\mathbf{x}^u_i})), \quad
	\widetilde{\mathbf{x}^u_i} = T_\mathsf{weak}(\mathbf{x}^u_i) \\
	& \widehat{p^i_{tea}} = \mathbf{M_\theta}\widetilde{p^i_{tea}}, \quad
	 \mathbf{M_\theta}=\small{\left[ \begin{array}{ccc} \cos(\theta)&\sin(\theta)&0 \\ -\sin(\theta)&\cos(\theta)&0 \\ 0&0&1 \end{array}\right]}
	\label{EqsRotCons}
\end{aligned}
\end{equation}
where $T_\mathsf{CutOcc}$ and $T_{\mathsf{Rot}_\theta}$ means our proposed strong augmentations CutOcc and in-plane rotation, respectively. $\mathbf{M_\theta}$ represents the in-plane rotation matrix corresponding to $\theta$. And $\widehat{p^i_{tea}}$ is the aligned prediction. We finally enforce consistency between distributions $\widehat{p^i_{tea}}$ and $\overline{p^i_{stu}}$.

It is worth noting that Kuhnke et al. \cite{kuhnke2021relative, kuhnke2023domain} also employed rotation-based consistency within semi-supervised HPE, but their methods were designed for constrained benchmarks (e.g., BIWI), whereas our proposed $T_\mathsf{CutOcc}$ and $T_{\mathsf{Rot}_\theta}$ are specifically tailored for unconstrained in-the-wild heads with severe occlusion and atypical orientations.

\subsection{Adaptation of SemiUHPE}

As discussed above, HPE is essentially a 3D rotation regression task. Therefore, SemiUHPE can be applied to any type of object without modification, and is roughly equivalent to a \textit{semi-supervised generic object rotation regression} (SemiObjRot)  framework. In experiments, we will quantitatively validate that SemiObjRot is superior to those specially designed methods \cite{yin2022fishermatch, yin2025towards} for regressing the rotation of common objects. Next, we explain how to simply adapt SemiUHPE so that it can be used to address the \textit{semi-supervised 3D head reconstruction} (Semi3DHead)  task.

\begin{figure}[t]
	\centering
	\includegraphics[width=\columnwidth]{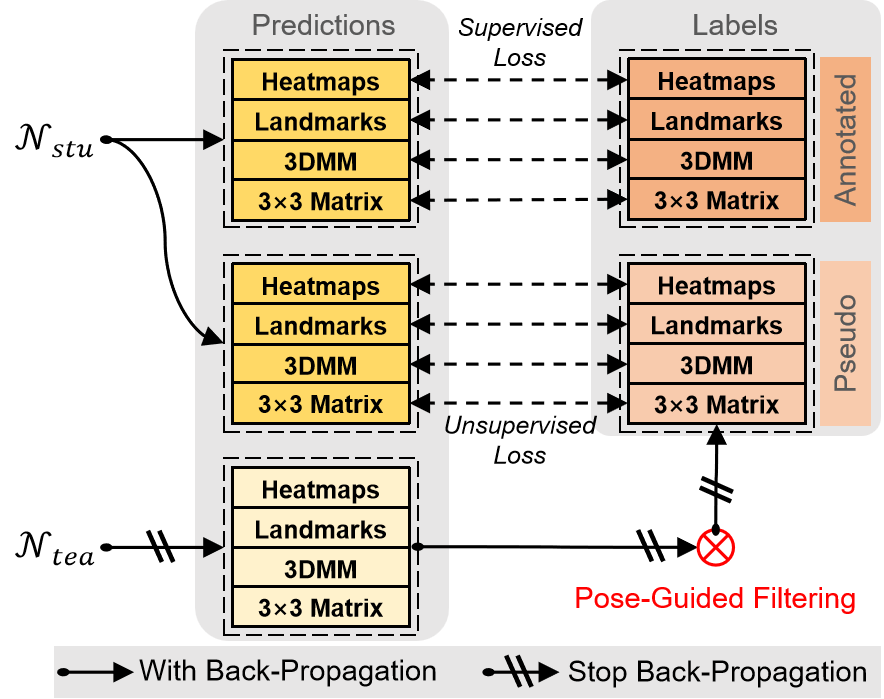}
	\vspace{-20pt}
	\caption{Overall architecture of our Semi3DHead framework. It extends DAD-3DNet \cite{martyniuk2022dad} by incorporating multiple input branches: (i) heatmaps and landmarks inherited from DAD-3DNet, (ii) a 3D Morphable Model (3DMM) branch using the FLAME model \cite{li2017learning}, and (iii) an additional $3\!\times\!3$ rotation matrix branch introduced in this work to directly regress head pose and filter out out-of-distribution samples.}
	\label{Framework2}
	\vspace{-10pt}
\end{figure}

Specifically, current mainstream 3D head reconstruction methods \cite{guo2020towards, sanyal2019learning, martyniuk2022dad} rely on the 3D Morphable Model (3DMM) (\eg, BFM \cite{paysan20093d} and FLAME \cite{li2017learning}) to transform it into a concise regression problem, which mainly predicts head shape and pose parameters. Then, motivated by the success of HPE using only monocular images \cite{yang2019fsa, hempel20226d, zhang2023tokenhpe}, we decide to add one additional HPE sub-branch on the basic reconstruction network. This branch can also be optimized smoothly for predicting 3D head pose (\eg, represented by the rotation matrix in $\mathcal{SO}(3)$), which is similar but independent of the head rotation in outputted 3DMM parameters. Without loss of generality,  as shown in Fig.~\ref{Framework2}, we choose DAD-3DNet \cite{martyniuk2022dad} as the basic main network and construct a modified semi-supervised version for 3D head reconstruction. It leverages small-scale labeled 3D heads and large-scale unlabeled wild heads to optimize the teacher-student mutual learning Mean-Teacher framework. Although DAD-3DNet has multiple output branches (such as 3DMM, heatmaps and landmarks), we can still follow the training idea of SemiUHPE. And we also experimentally confirmed that the newly added branch does not hinder the overall model performance.

\begin{figure}[t]
	\centering
	\includegraphics[width=\columnwidth]{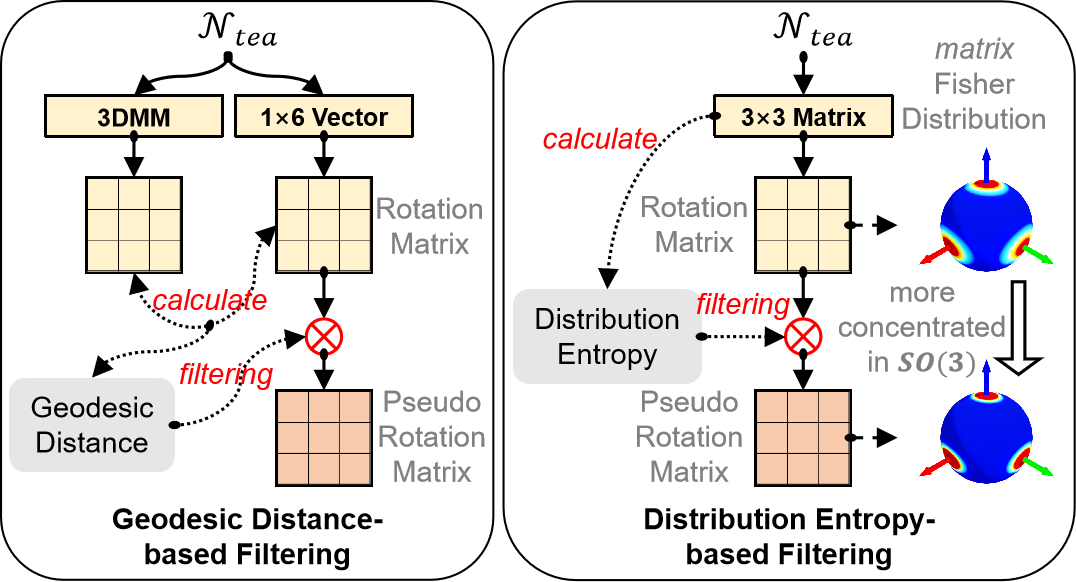}
	\vspace{-20pt}
	\caption{The illustration of two pose-guided pseudo-label filtering strategies based on geodesic distance (left) or distribution entropy (right).}
	\label{PoseFilter}
	\vspace{-10pt}
\end{figure}


To filter unreliable pseudo-labeled heads, we introduce a new pose regression branch and design a pose-guided filtering strategy. As illustrated in Fig.~\ref{PoseFilter} (left), one straightforward way is to measure the discrepancy between the predicted rotation matrix from the head-pose branch and the rotation matrix embedded in the 3DMM branch. This discrepancy can be quantified by the \textit{geodesic distance} on the rotation group $\mathcal{SO}(3)$, which corresponds to the minimal angular difference between two rotations (approximated by the Frobenius norm $\|\mathbf{I}\!-\!\mathbf{R}_1\mathbf{R}_2^T\|_F$). A smaller geodesic distance indicates higher consistency and reliability. However, relying solely on matrix differences may not always provide a statistically robust measure of uncertainty. To address this, we further model the head pose prediction probabilistically by fitting it with a matrix Fisher distribution, which naturally captures the distribution of rotations on $\mathcal{SO}(3)$. In this setting, the \textit{distribution entropy} serves as an indicator of prediction confidence: low entropy means the distribution is sharp and reliable, while high entropy corresponds to ambiguous or noisy samples. As shown in Fig.~\ref{PoseFilter} (right), we use this entropy-based criterion to dynamically filter out high-uncertainty samples. There are two advantages to applying the matrix Fisher distribution. On one hand, its output is an arbitrary $3\!\times\!3$ matrix with homeomorphic parameters to $\mathbb{R}^9$ \cite{mohlin2020probabilistic}, which is easier to estimate than the 3D rotations in $\mathcal{SO}(3)$ topology \cite{liu2021mfdnet}. On the other hand, we can measure the confidence of predicted distribution by calculating its entropy, and devise a self-reliant more robust filtering mechanism. Therefore, while different rotation representations can be mathematically transformed into each other, we adopt the $3\!\times\!3$ rotation matrix as the default representation. We will verify that the adapted SemiUHPE can benefit from both filtering strategies, with the latter being superior.

\subsection{Training Protocol}
Whether it is the original SemiUHPE or its adapted version, our training has two consecutive phases.

\textsf{Phase1:} We train the student model $\mathcal{N}_{stu}$ to learn a rotation regressor on labeled set $\mathcal{D}^l$ with a supervised loss:
\begin{equation}
	\mathcal{L}_{sup}(\mathbf{x}^l, \mathbf{y}^l) = -\log(\mathcal{MF}(\mathbf{y}^l; \mathcal{N}_{stu}(\mathbf{x}^l))
	\label{SupLoss}
\end{equation}
where $-\log(\cdot)$ means the negative log likelihood (NLL) of the mode predicted by $\mathcal{N}_{stu}$ in the distribution of ground-truth label $\mathbf{y}^l$. We save the best performed student model for cloning an identical teacher model for the next phase.

\textsf{Phase2:} After supervised training, we obtain a pair of teacher and student networks with the same initialization. Now, we begin the semi-supervised phase on both labeled set $\mathcal{D}^l$ and unlabeled set $\mathcal{D}^u$. The total loss is:
\begin{equation}
	\mathcal{L} = \mathcal{L}_{sup}(\mathbf{x}^l, \mathbf{y}^l) + \lambda \mathcal{L'}_{unsup}(\mathbf{x}^u)
	\label{TotalLoss}
\end{equation}
where $\mathcal{L}_{sup}$ and $\mathcal{L'}_{unsup}$ are calculated as in Eq.~\ref{SupLoss} and Eq.~\ref{UnsupLossNew}, respectively. The $\lambda$ is a weight of unsupervised loss, which is set to 1 in all experiments. Besides, we usually allocate different iterations for two phases based on the complexity of labeled $\mathcal{D}^l$ and unlabeled $\mathcal{D}^u$ datasets.

\section{Experiments}\label{exps}

\subsection{Datasets and Settings}

We introduce two kinds of datasets for the HPE task. 

\textbf{Labeled Datasets:}
we adopt the popular benchmark 300W-LP \cite{zhu2016face} which has 122,450 samples with half flipping as the train-set and AFLW2000 \cite{zhu2016face} as the val-set for comparing with the mainstream front-range HPE methods. We also utilize a recent 3D head reconstruction dataset DAD-3DHeads \cite{martyniuk2022dad} with three subsets (37,840 images in train-set, 4,312 images in val-set, and 2,746 images in test-set) for implementing the full-range unconstrained HPE.

\textbf{Unlabeled Datasets:}
we utilize COCO \cite{lin2014microsoft} with wild human heads as the unlabeled set for boosting both the front-range and full-range HPE. COCO does not have the head box label. Thus, we utilize its variation COCO-HumanParts \cite{yang2020hier} with labeled head boxes, and preprocess it as in BPJDet \cite{zhou2023body, zhou2024bpjdet} to generate COCOHead, which has about 74K samples after removing heads smaller than 30 pixels. These left heads cover diverse scenarios and cases. Besides, we also adopt this processing paradigm to extract valid head images from other human-related datasets including WiderFace \cite{yang2016wider}, CrowdHuman \cite{shao2018crowdhuman} and OpenImageV6 \cite{xie2020unsupervised}, which can be used for further ablation tests and model improvement in following subsections.

Then, we designed three increasingly difficult SSL settings along with a group of compared methods (\eg, both fully supervised and semi-supervised) as described below:

\textbf{Setting1: 300W-Self~}
In this setting, we use partially annotated 300W-LP as the labeled set and the left part as the unlabeled set. The labeled ratio is selected from (2\%, 5\%, 10\%, 20\%). The \textsf{Phase1} and \textsf{Phase2} has 20K and 40K iterations, respectively. The number of threshold updating stages $K$ in \textsf{Phase2} is 4. Batch size for the labeled set $B_l$ and unlabeled set $B_u$ is 32 and 128, respectively. The remained percentage $\delta$ of unlabeled data is 0.95. The test-set is AFLW2000. For a fair comparison, we keep samples with Euler angles within $\pm90^\circ$ following the previous front-range HPE methods \cite{yang2019fsa, zhou2020whenet, hempel20226d}. The evaluation metric is Mean Absolute Error (MAE) of Euler angles. We thus convert predictions into Euler angles for comparing.

\textbf{Setting2: 300W-COCOHead~}
Still for the front-range HPE, we combine all labeled 300W-LP with additional unlabeled faces in COCOHead for further performance boosting. We set \textsf{Phase1} and \textsf{Phase2} with 180K and 60K iterations, respectively. Parameters $K$, $B_l$, $B_u$ and $\delta$ are set to 6, 16, 128 and 0.75, respectively. The test-set is also AFLW2000. The others are the same as \textbf{Setting1}.

\textbf{Setting3: DAD-COCOHead~}
This is for the full-range HPE task. We use the train-set of DAD-3DHeads as the labeled set, and COCOHead as the unlabeled set. We set \textsf{Phase1} and \textsf{Phase2} with 100K and 100K iterations, respectively. And parameters $K$, $B_l$, $B_u$ and $\delta$ are set to 5, 16, 128 and 0.75, respectively. We report testing results on both val-set and test-set of DAD-3DHeads. Following \cite{martyniuk2022dad}, measures of the ground-truth matrix $\mathbf{R}_1$ and predicted $\mathbf{R}_2$ are (1) Frobenius norm of the matrix $\mathbf{I}\!-\!\mathbf{R}_1\mathbf{R}_2^T$, and (2) the angle in axis-angle representation of $\mathbf{R}_1\mathbf{R}_2^T$ (\textit{a.k.a} the geodesic distance between two rotation matrices).

\textbf{Compared Methods:}
In order to make a more comprehensive comparison, apart from our proposed SemiUHPE, we also implemented the fully supervised version (Sup.) using only labeled data but adopting an advanced matrix Fisher representation \cite{mohlin2020probabilistic}, and the semi-supervised baseline FisherMatch \cite{yin2022fishermatch} (Base.) following our settings. For the network backbone, it is selected from three purely ConvNets-based candidates including ResNet50 \cite{he2016deep}, RepVGG \cite{ding2021repvgg} and EfficientNetV2-S \cite{tan2021efficientnetv2}. These backbones have similar network parameters and are widely used in other HPE methods \cite{yang2019fsa, zhou2020whenet, hempel20226d}. Besides, we also included two additional baselines to strengthen the comparisons: (1) FisherMatch+ \cite{yin2025towards} (Base.+), where we replace the matrix Fisher distribution with a rotation Laplace distribution; (2) Sample Adaptive Augmentation \cite{gui2023enhancing} (SAA), an SSL method from image classification that introduces stronger adaptive augmentations. Both Base.+ and SAA were re-implemented on top of the best-performing settings of FisherMatch for SemiUHPE. For all experiments, we implemented them using PyTorch 1.13.1 on one single RTX 3090 (24 GB) or A800 (80 GB) GPU. The Adam optimizer is used. The learning rate in \textsf{Phase1} is 1e-4, and reduced to 1e-5 in \textsf{Phase2}. During training and testing, all head images are padded and resized into shape $224\times224$ as inputs.


\textbf{Adapted SemiUHPE:}
Similarly, we introduce the semi-supervised experimental data and settings for generic object rotation regression (SemiObjRot) and 3D head reconstruction (Semi3DHead). For SemiObjRot, we follow the setting in FisherMatch \cite{yin2022fishermatch} and conduct SSL comparing tests on the dataset Pascal3D+ \cite{xiang2014beyond}. This dataset contains real images from Pascal VOC and ImageNet of 12 rigid object classes. We evaluate 6 vehicle categories (aeroplane, bicycle, boat, bus, car, motorbike) which have relatively evenly distributed poses in azimuth angles, and set the number of labeled images as 7, 20 and 50 for each category respectively. For Semi3DHead, we conduct SSL comparing tests on the dataset DAD-3DHeads \cite{martyniuk2022dad}, where part of the train-set is used as labeled data (\eg, 5\%, 10\% or 20\%), the rest is used as unlabeled data, and the val-set is used to report performance. The comparison methods are roughly the same as SemiUHPE.

\subsection{Implementation Details}

Following above-defined three SSL settings, we provide many more details of implementing our proposed SemiUHPE and compared methods.

\textbf{Confidence Threshold in FisherMatch:}
In order to fairly compare SemiUHPE with the baseline method FisherMatch \cite{yin2022fishermatch}, we must select an appropriate pre-defined confidence threshold for it. In \textbf{Setting1}, we have determined the optimal threshold $\tau\!=\!-5.4$ by running several controlled experiments as shown in Fig.~\ref{asExpsA}. Then, in \textbf{Setting2}, we selected the final computed dynamic threshold $\tau_{6}\!=\!-4.3$ of SemiUHPE (with either ResNet50 \cite{he2016deep}, RepVGG \cite{ding2021repvgg} or EfficientNetV2-S \cite{tan2021efficientnetv2} as the backbone) as the pre-fixed threshold for FisherMatch. Similarly, in \textbf{Setting3}, we selected the final computed dynamic threshold $\tau_{5}\!=\!-4.8$, $\tau_{5}\!=\!-4.7$ and $\tau_{5}\!=\!-4.8$ of SemiUHPE with the backbone ResNet50, RepVGG and EfficientNetV2-S as the pre-fixed threshold, respectively. As shown in Table~\ref{tabOne}, Table~\ref{tabTwo} and Table~\ref{tabThree}, the performance of implemented FisherMatch is comparable or better than fully supervised methods, which explains the correctness and effectiveness of this strong baseline. Nonetheless, our SemiUHPE can always exceed FisherMatch by a clear margin.

\textbf{Rotation Consistency for dataset 300W-LP:}
For the pose-altering rotation consistency part $T_{\mathsf{Rot}_\theta}$ in our proposed head-oriented strong augmentation, we observed that it does not have a positive effect on semi-supervised HPE task related to the 300W-LP \cite{zhu2016face} dataset. The main factor causing this obstacle is that the 300W-LP dataset does not use an accurate $3\!\times\!3$ rotation matrix to represent the head pose label, but rather chooses the Euler angles with inherent flaws, including the \textit{gimbal lock} and \textit{discontinuity}. This further prevents us from aligning the two matrices after random rotation. Therefore, we did not apply $T_{\mathsf{Rot}_\theta}$ in \textbf{Setting1} or \textbf{Setting2}. This defect does not exist in \textbf{Setting3} that uses the labeled set DAD-3DHeads \cite{martyniuk2022dad} with the rotation matrix as its head pose label.

\setlength{\tabcolsep}{5pt}
\begin{table}[!t]  
	\begin{center}
	\caption{Hyper-parameters in paired weak-strong augmentations.}
	\vspace{-10pt}
	\begin{tabular}{cc|c|c|c}
	\Xhline{1.2pt}
	\multicolumn{2}{c|}{Is Used?} & \multirow{2}{*}{Aug.} & \multirow{2}{*}{Parameter} & \multirow{2}{*}{Probability} \\
	\cline{1-2}
	Weak & Strong & ~ & ~ & ~ \\
	\Xhline{1.2pt}
	\cmark & \cmark & Flip & Horizontally & 0.5 \\
	\cmark & \cmark & Blur & --- & 0.05 \\
	\cmark & \xmark & Scale & $s\in[0.8, 1.25]$ & 1.0 \\  
	\xmark & \cmark & Scale & $s\in[0.6, 1.5]$ & 1.0 \\  
	\xmark & \cmark & $T_\mathsf{CutOcc}$ & \makecell{CutOut: 3 Holes\\ CutMix: 3 Holes} & 1.0 \\
	\xmark & \cmark & $T_{\mathsf{Rot}_\theta}$ & $\theta\in(-30^\circ, 30^\circ)$ & 1.0 \\
	\Xhline{1.2pt}
	\end{tabular}
	\label{tabHyperParam}
	\end{center}
	\vspace{-10pt}
\end{table}

\textbf{Parameters of Augmentations:}
Hyper-parameters related to augmentations are shown in Table~\ref{tabHyperParam}. Specifically, Flip and Blur mean left-right flipping and simple image filter, respectively. Scale means the random resized crop. Both $T_\mathsf{CutOcc}$ and $T_{\mathsf{Rot}_\theta}$ are proposed by us. And the number of mask holes in CutOut \cite{devries2017improved} or CutMix \cite{yun2019cutmix} means generated random patches. Generally, it is laborious and impractical to undergo ablation tests for searching optimal values of these parameters. Therefore, these hyper-parameters are empirically initialized.

\textbf{SemiObjRot and Semi3DHead:}
For SemiObjRot, we inherit most of the training parameter settings in FisherMatch, except that the newly added parameter $\delta$ is set to 0.95 and strong augmentations are replaced with our proposed ones. Besides, we evaluate the experiments by the mean error, the median error (in degrees) and the accuracy within $30^\circ$ between the prediction and the ground truth. For Semi3DHead, we follow DAD-3DNet and \textbf{Setting1} to conduct SSL training experiments. The evaluation metrics are updated into Reprojection NME, Z$_n$ accuracy, Chamfer Distance and Pose Error for the 3D head learning task. More details can be found in \cite{yin2022fishermatch} and \cite{martyniuk2022dad}.

\subsection{Quantitative Comparison}
Below, we report results under three settings as well as improvement analysis of our SemiUHPE. If there is no special reminder, all baselines (Sup., Base., Base.+ and SAA) are implemented with our proposed aspect-ratio invariant cropping. The last is the results of SemiObjRot and Semi3DHead.

\setlength{\tabcolsep}{4pt}
\begin{table}[!t]  
	\begin{center}
	\caption{Euler angles errors on AFLW2000. Models are trained on 300W-LP with different ratios of label. The best result is in \textcolor{red}{red} color. SL and SSL are fully supervised learning and semi-supervised learning, respectively.}
	\vspace{-10pt}
	\begin{tabular}{c|l|c|ccccc}
	\Xhline{1.2pt}
	Type & Method & Backbone & 2\% & 5\% & 10\% & 20\% & All \\
	\Xhline{1.2pt}
	\multirow{3}{*}{SL} & Sup. & ResNet50 & 4.347 & 3.987 & 3.831 & 3.619 & 3.578 \\
	~ & Sup. & RepVGG & 4.252 & 3.817 & 3.724 & 3.579 & 3.498 \\
	~ & Sup. & EffNetV2-S & 4.009 & 3.678 & 3.517 & 3.444 & \textcolor{red}{3.379} \\
	\hline
	\multirow{8}{*}{SSL} & \tikzmark{bbw}Base. & ResNet50 & 4.190 & 3.798 & 3.609 & 3.492 & --- \\
	~ & \tikzmark{ddw}Base. & RepVGG & 4.023 & 3.730 & 3.489 & 3.445 & --- \\ 
	~ & \tikzmark{ffw}Base. & EffNetV2-S & 3.991 & 3.596 & 3.448 & 3.372 & --- \\
	\cline{2-8}
	~ & Base.+ & ResNet50 & 4.102 & 3.702 & 3.514 & 3.475 & --- \\
	~ & SAA & ResNet50 & 4.038 & 3.654 & 3.508 & 3.469 & --- \\
	\cline{2-8}
	~ & \tikzmark{aaw}Ours & ResNet50 & 3.956 & 3.629 & 3.487 & 3.463 & --- \\
	~ & \tikzmark{ccw}Ours & RepVGG & 3.953 & 3.607 & 3.510 & 3.424 & --- \\
	~ & \tikzmark{eew}Ours & EffNetV2-S & \textcolor{red}{3.835} & \textcolor{red}{3.526} & \textcolor{red}{3.377} & \textcolor{red}{3.348} & --- \\
	\Xhline{1.2pt}
	\end{tabular}
    	\begin{tikzpicture}[ remember picture, overlay, thick]
   	 	\draw [<-,green] ([xshift=-0.5ex, yshift=1ex]pic cs:bbw) [bend right] to ([xshift=-0.5ex, yshift=1ex]pic cs:aaw);
    		\draw [<-,magenta] ([xshift=-0.5ex, yshift=1ex]pic cs:ddw) [bend right] to ([xshift=-0.5ex, yshift=1ex]pic cs:ccw);
		\draw [<-,cyan] ([xshift=-0.5ex, yshift=1ex]pic cs:ffw) [bend right] to ([xshift=-0.5ex, yshift=1ex]pic cs:eew);
	\end{tikzpicture}
	\label{tabOne}
	\end{center}
	\vspace{-10pt}
\end{table}

\setlength{\tabcolsep}{1.5pt}
\begin{table}[!t]  
	\begin{center}
	\caption{Euler angles errors on AFLW2000. Models are all trained on 300W-LP. Extra means additional annotations (\eg, Landmarks (LMs) or 3DMM). The marker $\dag$ indicates using additional labeled training data. The best and second-best result is in \textcolor{red}{red} and \textcolor{blue}{blue} color, respectively.}
	\vspace{-10pt}
	\begin{tabular}{c|l|c|c|cccc}
	\Xhline{1.2pt}
	Type & Method & Reference & Extra & Pitch & Yaw & Roll & MAE \\
	\Xhline{1.2pt}
	\multirow{25}{*}{SL} & 3DDFA \cite{zhu2016face} & CVPR'16 & 3DMM & 5.98 & 4.33 & 4.30 & 4.87 \\
	~ & HopeNet \cite{ruiz2018fine} & CVPRW'18 & No & 6.56 & 6.47 & 5.44 & 6.16 \\
	~ & QuatNet \cite{hsu2018quatnet} & TMM'18 & No & 5.62 & 3.97 & 3.92 & 4.50 \\
	~ & FSA-Net \cite{yang2019fsa} & CVPR'19 & No & 6.08 & 4.50 & 4.64 & 5.07 \\
	~ & WHENet-V \cite{zhou2020whenet} & BMVC'20 & No & 5.75 & 4.44 & 4.31 & 4.83 \\
	~ & FDN \cite{zhang2020fdn} & AAAI'20 & No & 5.61 & 3.78 & 3.88 & 4.42 \\
	~ & 3DDFA-V2 \cite{guo2020towards} & ECCV'20 & 3DMM & 5.26 & 4.06 & 3.48 & 4.27 \\
	~ & MNN \cite{valle2020multi} & TPAMI'20 & LMs & 4.69 & 3.34 & 3.48 & 3.83 \\
	~ & Rankpose \cite{dai2020rankpose} & BMVC'20 & No & 4.75 & 2.99 & 3.25 & 3.66 \\
	~ & TriNet \cite{cao2021vector} & WACV'21 & No & 5.77 & 4.20 & 4.04 & 4.67 \\
	~ & MFDNet \cite{liu2021mfdnet} & TMM'21 & No & 5.16 & 4.30 & 3.69 & 4.38 \\
	~ & Img2Pose \cite{albiero2021img2pose} & CVPR'21 & LMs & 5.03 & 3.43 & 3.28 & 3.91 \\
	~ & SADRNet \cite{ruan2021sadrnet}$\dag$ & TIP'21 & 3DMM & 5.00 & 2.93 & 3.54 & 3.82 \\
	~ & SynergyNet \cite{wu2021synergy} & 3DV'21 & 3DMM & \textcolor{red}{4.09} & 3.42 & \textcolor{red}{2.55} & 3.35 \\
	~ & 6DRepNet \cite{hempel20226d} & ICIP'22 & No & 4.91 & 3.63 & 3.37 & 3.97 \\
	~ & DAD-3DNet \cite{martyniuk2022dad}$\dag$ & CVPR'22 & 3DMM & 4.76 & 3.08 & 3.15 & 3.66 \\
	~ & TokenHPE \cite{zhang2023tokenhpe} & CVPR'23 & No & 5.54 & 4.36 & 4.08 & 4.66 \\    
	~ & DSFNet-f \cite{li2023dsfnet}$\dag$ & CVPR'23 & 3DMM & \textcolor{blue}{4.28} & \textcolor{red}{2.65} & 2.82 & \textcolor{red}{3.25} \\
	~ & CIT-v1 \cite{li2023cascaded} & IJCV'23 & LMs & 4.38 & \textcolor{blue}{2.68} & 3.45 & 3.50 \\
	~ & 2DHeadPose \cite{wang20232dheadpose}$\dag$ & NN'23 & No & 4.47 & 2.85 & 2.82 & 3.38 \\
	~ & HeadDiff \cite{wang2024headdiff} & TIP'24 & No & 4.55 & 3.15 & 3.03 & 3.57 \\
	~ & OPAL (6D) \cite{cobo2024representation}  & PR'24 & No & 4.59 & 2.85 & 3.04 & 3.49 \\
	\cline{2-8}
	~ & Sup. (ResNet50) & --- & No & 4.58 & 3.20 & 2.95 & 3.58 \\
	~ & Sup. (RepVGG) & --- & No & 4.50 & 3.18 & 2.81 & 3.50 \\
	~ & Sup. (EffNetV2-S) & --- & No & 4.40 & 2.99 & 2.75 & 3.38 \\
	\hline
	\multirow{8}{*}{SSL} & \tikzmark{bbx}Base. (ResNet50) & CVPR'22 & No & 4.61 & 2.99 & 3.00 & 3.53 \\
	~ & \tikzmark{ddx}Base. (RepVGG) & CVPR'22 & No & 4.46 & 2.84 & 2.77 & 3.36 \\
	~ & \tikzmark{ffx}Base. (EffNetV2-S) & CVPR'22 & No & 4.39 & 2.87 & 2.79 & 3.35 \\
	\cline{2-8}
	~ & Base.+ (ResNet50) & TPAMI'25 & No & 4.58 & 2.96 & 2.83 & 3.46 \\
	~ & SAA (ResNet50) & CVPR'23 & No & 4.56 & 2.96 & 2.79 & 3.44 \\
	\cline{2-8}
	~ & \tikzmark{aax}Ours (ResNet50) & --- & No & 4.52 & 2.89 & \textcolor{blue}{2.71} & 3.37 \\
	~ & \tikzmark{ccx}Ours (RepVGG) & --- & No & 4.43 & 2.86 & 2.75 & 3.35 \\
	~ & \tikzmark{eex}Ours (EffNetV2-S) & --- & No & 4.39 & 2.79 & 2.77 & \textcolor{blue}{3.31} \\
	\Xhline{1.2pt}
	\end{tabular}
    	\begin{tikzpicture}[ remember picture, overlay, thick]
   	 	\draw [<-,green] ([xshift=-0.5ex, yshift=1ex]pic cs:bbx) [bend right] to ([xshift=-0.5ex, yshift=1ex]pic cs:aax);
    		\draw [<-,magenta] ([xshift=-0.5ex, yshift=1ex]pic cs:ddx) [bend right] to ([xshift=-0.5ex, yshift=1ex]pic cs:ccx);
		\draw [<-,cyan] ([xshift=-0.5ex, yshift=1ex]pic cs:ffx) [bend right] to ([xshift=-0.5ex, yshift=1ex]pic cs:eex);
	\end{tikzpicture}
	\label{tabTwo}
	\end{center}
	\vspace{-10pt}
\end{table}

\setlength{\tabcolsep}{1.5pt}
\begin{table}[!t]  
	\begin{center}
	\caption{HPE results on the \textit{val-set} and \textit{test-set} of challenging DAD-3DHeads dataset. The marker * indicates additional \textit{ublabeled} data is used.}
	\vspace{-10pt}
	\begin{tabular}{c|l|c|c|c|c}
	\Xhline{1.2pt}
	\multirow{2}{*}{Type} & \multirow{2}{*}{Method} & \multicolumn{2}{c|}{$\|\mathbf{I}\!-\!\mathbf{R}_1\mathbf{R}_2^T\|_F\!\downarrow$} & \multicolumn{2}{c}{Angle error (degree)$\downarrow$} \\
	\cline{3-6}
	~ & ~ & \; val-set \; & \; test-set \; & \; val-set \; & test-set \\
	\Xhline{1.2pt}
	\multirow{5}{*}{SL} & Img2Pose \cite{albiero2021img2pose} & --- & 0.226 & --- & 9.122 \\
	~ & DAD-3DNet \cite{martyniuk2022dad} & 0.130 & 0.138 & 5.456 & 5.360 \\
	\cline{2-6}
	~ & Sup. (ResNet50) & 0.133 & 0.138 & 5.543 & 5.234 \\
	~ & Sup. (RepVGG) & 0.128 & 0.134 & 5.321 & 5.020 \\
	~ & Sup. (EffNetV2-S) & 0.121 & \textcolor{blue}{0.125} & 4.993 & 4.728 \\
	\hline
	\multirow{7}{*}{SSL} & \tikzmark{bby}Base. (ResNet50) & 0.138 & 0.145 & 5.794 & 5.312 \\
	~ & \tikzmark{ddy}Base. (RepVGG) & 0.130 & 0.137 & 5.425 & 5.182 \\
	~ & \tikzmark{ffy}Base. (EffNetV2-S) & 0.121 & 0.127 & 5.013 & 4.799 \\
	~ & \tikzmark{aay}Ours (ResNet50) & 0.116 & 0.127 & 4.800 & 4.810 \\
	~ & \tikzmark{ccy}Ours (RepVGG) & 0.116 & 0.126 & 4.778 & 4.760 \\
	~ & \tikzmark{eey}Ours (EffNetV2-S) & \textcolor{blue}{0.112} & \textcolor{red}{0.124} & \textcolor{blue}{4.636} & \textcolor{blue}{4.636} \\
	\cline{2-6}
	~ & Ours (EffNetV2-S)* & \textcolor{red}{0.109} & \textcolor{red}{0.124} & \textcolor{red}{4.518} & \textcolor{red}{4.408} \\
	\Xhline{1.2pt}
	\end{tabular}
    	\begin{tikzpicture}[ remember picture, overlay, thick]
   	 	\draw [<-,green] ([xshift=-0.5ex, yshift=1ex]pic cs:bby) [bend right] to ([xshift=-0.5ex, yshift=1ex]pic cs:aay);
    		\draw [<-,magenta] ([xshift=-0.5ex, yshift=1ex]pic cs:ddy) [bend right] to ([xshift=-0.5ex, yshift=1ex]pic cs:ccy);
		\draw [<-,cyan] ([xshift=-0.5ex, yshift=1ex]pic cs:ffy) [bend right] to ([xshift=-0.5ex, yshift=1ex]pic cs:eey);
	\end{tikzpicture}
	\label{tabThree}
	\end{center}
	\vspace{-10pt}
\end{table}

\setlength{\tabcolsep}{12pt}
\begin{table*}[!t]  
	\begin{center}
	\caption{Detailed HPE errors on the \textit{test-set} of DAD-3DHeads with four subsets: challenging atypical poses (Pose), compound expressions (Expr.), heavy occlusions (Occl.) and non-standard light (Light). All of our results are returned after the prediction results are submitted to the official and compared with the undisclosed ground-truth labels. The best and second-best result is in \textcolor{red}{red} and \textcolor{blue}{blue} color, respectively.}
	\vspace{-10pt}
	\begin{tabular}{c|l|c|c|c|c|c}
	\Xhline{1.2pt}
	\multirow{2}{*}{Type} & \multirow{2}{*}{Method} & \multicolumn{5}{c}{$\|\mathbf{I}\!-\!\mathbf{R}_1\mathbf{R}_2^T\|_F\!\downarrow$ \;\;/\;\; Angle error (degree)$\downarrow$} \\
	\cline{3-7}
	~ & ~ & Overall & Pose & Expr. & Occl. & Light \\
	\Xhline{1.2pt}
	\multirow{6}{*}{SL} & 3DDFA-V2 \cite{guo2020towards} & 0.527 / \;\;-----\; & 0.790 / \;\;-----\;\; & 0.455 / \;\;-----\; & 0.542 / \;\;-----\; & \;\;-----\; / \;\;-----\; \\ 
	~ & RingNet \cite{sanyal2019learning} & 0.438 / \;\;-----\; & 1.076 / \;\;-----\;\; & 0.294 / \;\;-----\; & 0.551 / \;\;-----\; & \;\;-----\; / \;\;-----\; \\
	~ & DAD-3DNet \cite{martyniuk2022dad} & 0.138 / 5.360 & 0.343 / \;\;-----\;\; & 0.112 / \;\;-----\; & 0.203 / \;\;-----\; & \;\;-----\; / \;\;-----\; \\
	\cline{2-7}
	~ & Sup. (ResNet50) & 0.138 / 5.234 & 0.327 / 10.782 & 0.111 / 4.351 & 0.181 / 7.134 & 0.129 / 5.094 \\
	~ & Sup. (RepVGG) & 0.134 / 5.020 & 0.325 / \;\;9.961 & 0.108 / 4.233 & 0.179 / 6.880 & 0.131 / 4.904 \\
	~ & Sup. (EffNetV2-S) & \textcolor{blue}{0.125} / 4.728 & \textcolor{red}{0.274} / \;\;9.055 & 0.105 / 4.105 & 0.170 / 6.449 & 0.123 / 4.969 \\
	\hline
	\multirow{7}{*}{SSL} & \tikzmark{bbz}Base. (ResNet50) & 0.145 / 5.312 & 0.400 / 11.109 & 0.111 / 4.356 & 0.194 / 7.583 & 0.138 / 5.523 \\
	~ & \tikzmark{ddz}Base. (RepVGG) & 0.137 / 5.182 & 0.349 / 10.932 & 0.110 / 4.305 & 0.176 / 7.175 & 0.131 / 5.080 \\
	~ & \tikzmark{ffz}Base. (EffNetV2-S) & 0.127 / 4.799 & \textcolor{blue}{0.278} / \;\;9.244 & 0.106 / 4.135 & 0.178 / 6.372 & 0.122 / 4.864 \\ 
	~ & \tikzmark{aaz}Ours (ResNet50) & 0.127 / 4.810 & 0.322 / \;\;9.581 & 0.103 / 4.106 & 0.159 / 6.463 & 0.127 / 4.974 \\
	~ & \tikzmark{ccz}Ours (RepVGG) & 0.126 / 4.760 & 0.307 / \;\;9.110 & 0.105 / 4.180 & 0.149 / 6.041 & \textcolor{blue}{0.121} / \textcolor{blue}{4.683} \\
	~ & \tikzmark{eez}Ours (EffNetV2-S) & \textcolor{red}{0.124} / \textcolor{blue}{4.636}& 0.309 / \textcolor{blue}{\;\;8.620} & \textcolor{blue}{0.102} / \textcolor{blue}{4.047} & \textcolor{blue}{0.145} / \textcolor{blue}{5.897} & 0.123 / 4.761 \\
	\cline{2-7}
	~ & Ours (EffNetV2-S)* & \textcolor{red}{0.124} / \textcolor{red}{4.408} & 0.349 / \textcolor{red}{\;\;7.666} & \textcolor{red}{0.098} / \textcolor{red}{3.916} & \textcolor{red}{0.132} / \textcolor{red}{5.362} & \textcolor{red}{0.116} / \textcolor{red}{4.434} \\
	\Xhline{1.2pt}
	\end{tabular}
    	\begin{tikzpicture}[ remember picture, overlay, thick]
   	 	\draw [<-,green] ([xshift=-0.5ex, yshift=1ex]pic cs:bbz) [bend right] to ([xshift=-0.5ex, yshift=1ex]pic cs:aaz);
    		\draw [<-,magenta] ([xshift=-0.5ex, yshift=1ex]pic cs:ddz) [bend right] to ([xshift=-0.5ex, yshift=1ex]pic cs:ccz);
		\draw [<-,cyan] ([xshift=-0.5ex, yshift=1ex]pic cs:ffz) [bend right] to ([xshift=-0.5ex, yshift=1ex]pic cs:eez);
	\end{tikzpicture}
	\label{tabThreePlus}
	\end{center}
	\vspace{-10pt}
\end{table*}

\begin{figure*}[t]
	\centering
	\subfloat[]{\includegraphics[width=0.43\textwidth]{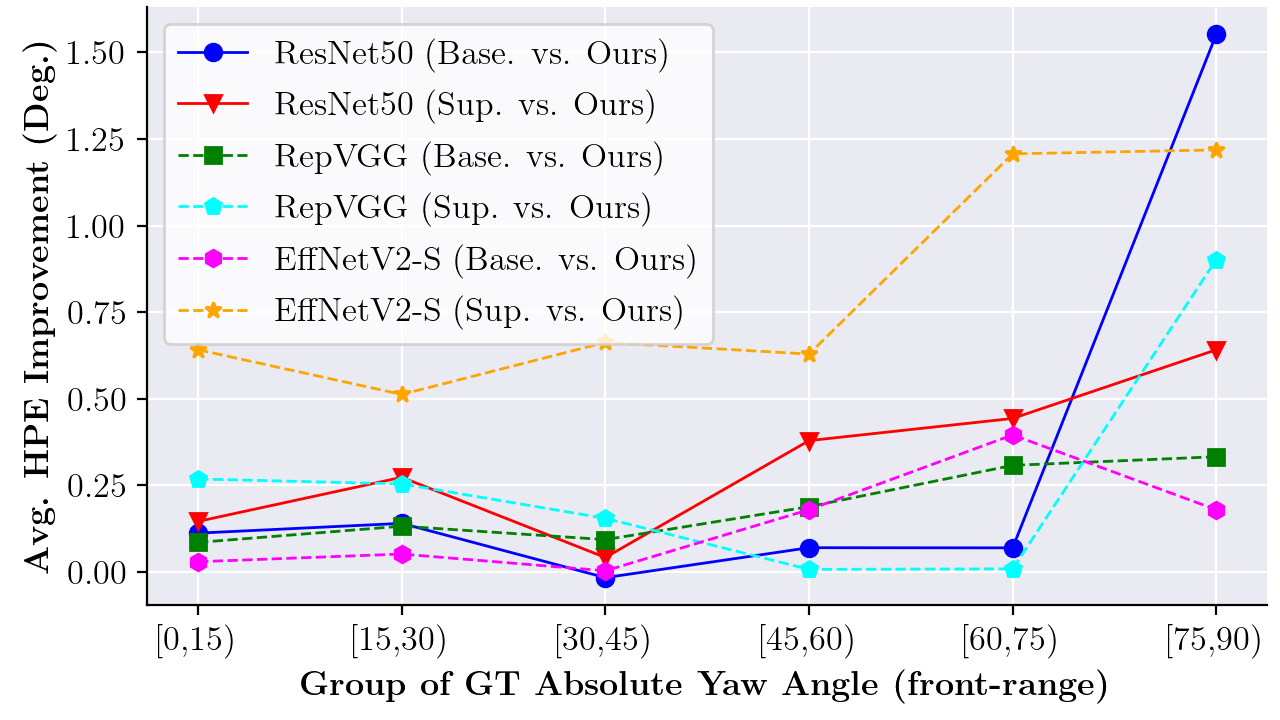}
	\label{impAnalysisA}}
	\subfloat[]{\includegraphics[width=0.56\textwidth]{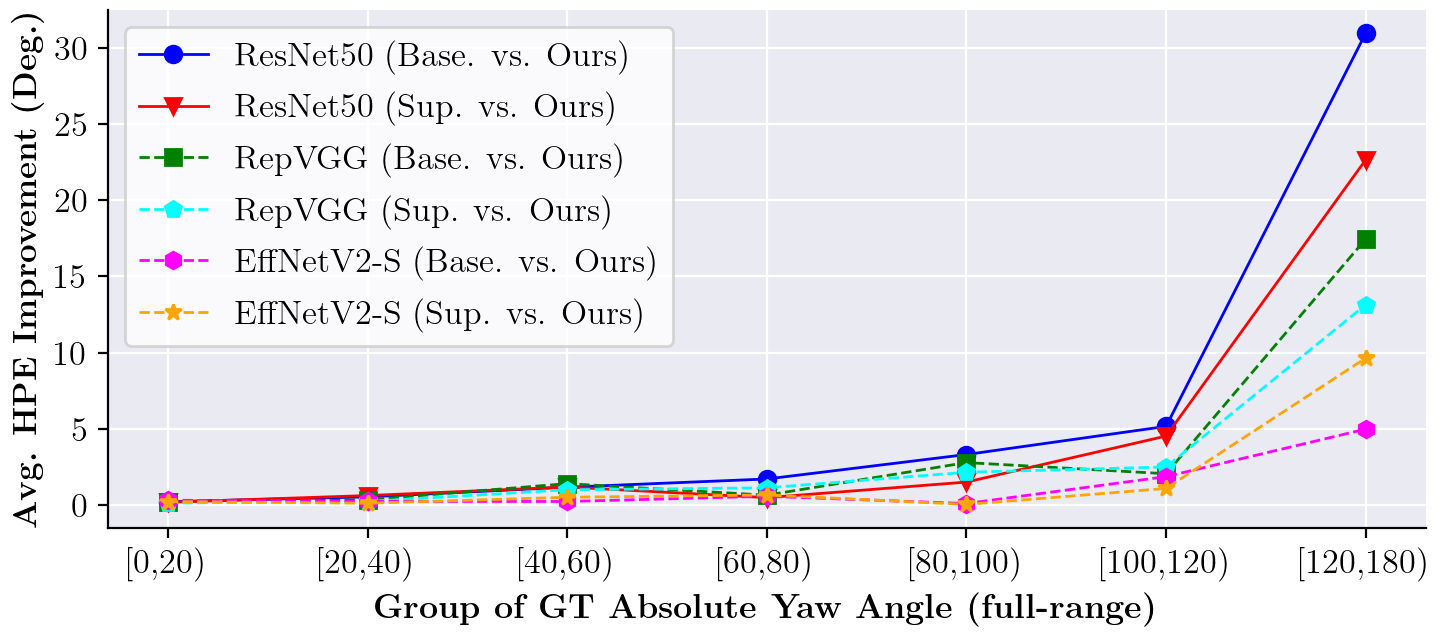}
	\label{impAnalysisB}}
	\vspace{-10pt}
	\caption{The average HPE improvement on (a) AFLW2000 under \textbf{Setting2} and (b) DAD-3DHeads val-set under \textbf{Setting3}. }
	\label{impAnalysis}
	\vspace{-10pt}
\end{figure*}

\setlength{\tabcolsep}{1.5pt}
\begin{table}[!t]  
	\begin{center}
	\caption{Comparing our method with the baselines on the 6 categories of Pascal3D+ dataset with few annotations. All results are averaged.}
	\vspace{-10pt}
	\begin{tabular}{l|cc|cc|cc}
	\Xhline{1.2pt}
	\multirow{2}{*}{Method} & \multicolumn{2}{c|}{7 images} & \multicolumn{2}{c|}{20 images} & \multicolumn{2}{c}{50 images} \\
	\cline{2-7}
	~ & Med.$\downarrow$ & Acc$_{30^\circ}$$\uparrow$ & Med.$\downarrow$ & Acc$_{30^\circ}$$\uparrow$ & Med.$\downarrow$ & Acc$_{30^\circ}$$\uparrow$ \\
	\Xhline{1.2pt}
	Res50-Gene & 39.1 & 36.1 & 26.3 & 45.2 & 20.2 & 54.6 \\
	Res50-Spec & 46.5 & 29.6 & 29.4 & 42.8 & 23.0 & 50.4 \\
	StarMap \cite{zhou2018starmap} & 49.6 & 30.7 & 46.4 & 35.6 & 27.9 & 53.8\\
	NeMo \cite{wang2021nemo} & 60.0 & 38.4 & 33.3 & 51.7 & 22.1 & 69.3 \\
	NVSM  \cite{wang2021neural} & 37.5 & 53.8 & 28.7 & 61.7 & 24.2 & 65.6 \\
	FisherMatch \cite{yin2022fishermatch} & 28.3 & 56.8 & 23.8 & 63.6 & 16.1 & 75.7 \\
	SAA \cite{gui2023enhancing} & 26.0 & 57.4 & 22.0 & 64.7 & 15.2 & 78.1 \\  
	FisherMatch+ \cite{yin2025towards} & 25.5 & 58.0 & 21.4 & 65.1 & 14.8 & 78.9 \\  
	Ours (Res50) & \textcolor{red}{21.5} & \textcolor{red}{61.2} & \textcolor{red}{18.6} & \textcolor{red}{67.7} & \textcolor{red}{12.8} & \textcolor{red}{83.1} \\
	\hline
	Full Sup. & 8.1 & 89.6 & 8.1 & 89.6 & 8.1 & 89.6 \\
	\Xhline{1.2pt}
	\end{tabular}
	\label{tabSemiObjRot}
	\end{center}
	\vspace{-10pt}
\end{table}

\setlength{\tabcolsep}{2.5pt}
\begin{table*}[!t] 
	\begin{center}
	\caption{Comparison of  \textbf{3D head reconstruction} results on DAD-3DHeads \textit{val-set}. The backbone of all models is MobileNet-w1 \cite{howard2017mobilenets}. All four metrics are adopted from the paper \cite{martyniuk2022dad}. Baselines Mean-Teacher \cite{tarvainen2017mean} and FixMatch \cite{sohn2020fixmatch} used the original DAD-3DNet without adding the HPE sub-branch. The ratio 5\% (1,892 labels), 10\% (3,784 labels), 20\% (7,568 labels) or 100\% (37,840 labels) means labeled samples in DAD-3DHeads \textit{train-set}.}
	\label{tabSemi3DHead}
	\vspace{-10pt}
	\begin{tabular}{c|l|cccc|cccc|cccc|cccc}
	\Xhline{1.2pt}
	\multirow{2}{*}{\textbf{Type}} & \multirow{2}{*}{\textbf{Method}} & \multicolumn{4}{c|}{\textbf{NME$\downarrow$}} & \multicolumn{4}{c|}{\textbf{$Z_5$ Accuracy$\uparrow$}} & \multicolumn{4}{c|}{\textbf{Chamfer Distance$\downarrow$}} & \multicolumn{4}{c}{\textbf{Pose Error$\downarrow$}} \\
	\cline{3-18}
	~ & ~ & 5\% & 10\% & 20\% & 100\% & 5\% & 10\% & 20\% & 100\% & 5\% & 10\% & 20\% & 100\% & 5\% & 10\% & 20\% & 100\% \\
	\Xhline{1.2pt}
	SL & DAD-3DNet \cite{martyniuk2022dad} & 7.861 & 4.766 & 4.025 & 2.467 & 0.919 & 0.929 & 0.938 & 0.954 & 3.957 & 3.468 & 3.247 & 2.982 & 0.327 & 0.270 & 0.233 & 0.149 \\
	\hline
	\multirow{5}{*}{SSL} & Mean-Teacher \cite{tarvainen2017mean} & 5.599 & 4.969 & 4.860 & --- & 0.926 & 0.928 & 0.936 & --- & 3.459 & 3.320 & 3.383 & --- & 0.291 & 0.277 & 0.241 & --- \\
	~ & Mean-Teacher \cite{tarvainen2017mean} + $T_\mathsf{CutOcc}$ & 4.071 & 3.845 & 3.645 & --- & 0.936 & \textcolor{blue}{0.939} & 0.939 & --- & 3.267 & 3.210 & 3.084 & --- & 0.245 & 0.233 & 0.226 & --- \\
	~ & FixMatch \cite{sohn2020fixmatch} + Fixed Thre. & 4.460 & 3.790 & \textcolor{blue}{3.543} & --- & 0.935 & 0.938 & 0.939 & --- & 3.248 & \textcolor{blue}{3.127} & 3.075 & --- & 0.252 & 0.230 & 0.225 & --- \\
	~ & Ours (Geo. Dist. Filtering)  & \textcolor{blue}{3.990} & \textcolor{blue}{3.781} & 3.553 & --- & \textcolor{blue}{0.937} & \textcolor{blue}{0.939} & \textcolor{blue}{0.942} & --- & \textcolor{blue}{3.179} & 3.144 & \textcolor{blue}{3.046} & --- & \textcolor{blue}{0.239} & \textcolor{blue}{0.229} & \textcolor{blue}{0.221} & --- \\
	~ & Ours (Entropy Filtering) & \textcolor{red}{3.833} & \textcolor{red}{3.748} & \textcolor{red}{3.490} & --- & \textcolor{red}{0.941} & \textcolor{red}{0.941} & \textcolor{red}{0.947} & --- & \textcolor{red}{3.152} & \textcolor{red}{3.055} & \textcolor{red}{3.027} & --- & \textcolor{red}{0.228} & \textcolor{red}{0.217} & \textcolor{red}{0.206} & --- \\
	\Xhline{1.2pt}
	\end{tabular}
	\end{center}
	\vspace{-10pt}
\end{table*}

\textbf{Front-range HPE Results under Setting1.}
As shown in Table~\ref{tabOne}, our SemiUHPE can surpass both the supervised methods \cite{mohlin2020probabilistic} and baseline FisherMatch \cite{yin2022fishermatch} with either backbone. When using ResNet50, our SemiUHPE is always the best out of other three SSL methods (Base., Base.+ and SSA). Meanwhile, the smaller the ratio of labeled samples used (from 20\% to 2\%), the more significant the reduction in MAE errors. Moreover, with using 20\% labeled 300W-LP, the performance of semi-supervised baseline and our SemiUHPE can always exceed the supervised method using all labeled 300W-LP. These results are also comparable to some SOTA supervised methods in Table~\ref{tabTwo}. We attribute this to the stronger robustness brought by partially integrated unsupervised training of unlabeled data. It also indicates that pure supervised learning may cause final models overfitting on the train-set, while SSL can improve the adaptability and generalization to a certain extent.

\textbf{Front-range HPE Results under Setting2.}
As shown in Table~\ref{tabTwo}, with the supporting of SSL, our method based on ResNet50 can achieve a low MAE result \textbf{3.37}, which is comparable to other supervised SOTA front-range HPE methods such as DAD-3DNet \cite{martyniuk2022dad} with MAE \textbf{3.66}, SynergyNet \cite{wu2021synergy} with MAE \textbf{3.35} and DSFNet-f \cite{li2023dsfnet} with MAE \textbf{3.25}. Please note that all of them are 3DMM-based which require dense face landmark labels and complex 3D face reconstruction pipelines. While, our SemiUHPE only requires reasonable excavation of unlabeled wild heads, which is more practical and scalable in real applications. Comparing to other three SSL baselines (Base., Base.+ and SSA), our method still outperforms them under this full-range setting. When using a stronger backbone such as RepVGG as in 6DRepNet \cite{hempel20226d}, our method can obtain a lower result with MAE \textbf{3.35}, which further explains its generality and advantage. In particular, after adopting the superior backbone EfficientNetV2-S, we can further reduce the MAE errors into \textbf{3.31}, which achieved the second best performance so far.

\textbf{Full-range HPE Results under Setting3.}
As shown in Table~\ref{tabThree}, results of Img2Pose \cite{albiero2021img2pose} and DAD-3DNet \cite{martyniuk2022dad} are obtained from the paper \cite{martyniuk2022dad}, except for results of DAD-3DNet on the val-set which are evaluated by us using its official model. The supervised method is implemented on DAD-3DHeads train-set. Generally, our SemiUHPE significantly surpasses the compared DAD-3DNet and retrained baselines without using any 3D information of face or head, which again proves its superiority and versatility. By expanding the unsupervised dataset (\eg, only COCOHead) to a larger one, which is the combination of COCOHead, CrowdHuman and OpenImageV6 with totally about 403K heads, our method (with marker *) finally achieved lower pose estimation errors on both val-set and test-set. More than that, we give detailed comparison under different conditions in Table~\ref{tabThreePlus}. Our method is still clearly ahead in many challenging cases such as atypical poses, compound expressions, heavy occlusions and non-strandard light. These are common scenes of in-the-wild head images, which are less-touched by the mainstream HPE studies \cite{yang2019fsa, hempel20226d, wang2024headdiff} and benchmarks \cite{zhu2016face, fanelli2013random}. Without bells and whistles, we achieved uncontroversial SOTA results on the full-range HPE dataset DAD-3DHeads. And all visualization results shown in this paper are estimated based on the optimal model obtained under this setting.

\textbf{Performance Improvement Analysis of SemiUHPE.}
To systematically understand the source of improvement after boosting FisherMatch with our proposed strategies, we further calculate and plot the head pose error gains on datasets AFLW2000 for \textit{front-range} HPE under \textbf{Setting2} (Fig.~\ref{impAnalysisA}) and DAD-3DHeads for \textit{full-range} HPE under \textbf{Setting3} (Fig.~\ref{impAnalysisB}). We split all the testing images into different groups according to their GT yaw angles and calculate the improved error degree within each group. We can find that the improvement by our SemiUHPE gets more obvious as the head pose becomes more challenging and has a larger yaw angle (e.g., $>90^\circ$). This capability benefits from unsupervised joint training using a large number of human heads collected in any orientation.

\textbf{Results of SemiObjRot.}
The experiment results on Pascal3D+ dataset of generic object rotation regression are shown in Table~\ref{tabSemiObjRot}. The results illustrate that, with using the effective teacher-student mutual learning framework as well as our proposed SSL strategies (including the dynamic entropy-based pseudo label filtering and two advanced strong augmentations), our SemiObjRot significantly outperforms various baselines and three state-of-the-art methods (including FisherMatch, FisherMatch+, and the adapted SAA) under all different numbers of labeled images. This is strong evidence of the undisputed superiority and valuable versatility of our approach.

\textbf{Results of Semi3DHead.}
The results of our modified Semi3DHead on the 3D head reconstruction task compared with other baselines are summarized in Table~\ref{tabSemi3DHead}. Although 3D head reconstruction seems to be a completely different task, it has an intrinsic connection and synergy with HPE \cite{wu2021synergy}. Therefore, after adding the HPE prediction branch, our Semi3DHead always outperforms the other two baselines (Mean-Teacher \cite{tarvainen2017mean} and FixMatch \cite{sohn2020fixmatch}) regardless of the pseudo-label filtering scheme used. In addition, by observing the Mean-Teacher after applying the strong augmentation $T_\mathsf{CutOcc}$ (even some metrics are better than FixMatch based on pseudo-label filtering), we can see the universality of $T_\mathsf{CutOcc}$. The two pseudo-label filtering methods we designed are better than the FixMatch based on static thresholds, indicating that reasonable dynamic filtering strategies are universally applicable. At the same time, the entropy-based one using the matrix Fisher distribution is significantly better, which is consistent with the previous analysis. In the future, we can try to extend Semi3DHead to more 3D head reconstruction datasets to fully leverage its ability to mine unlabeled head images.

\begin{figure}[t]
	\centering
	\includegraphics[width=\columnwidth]{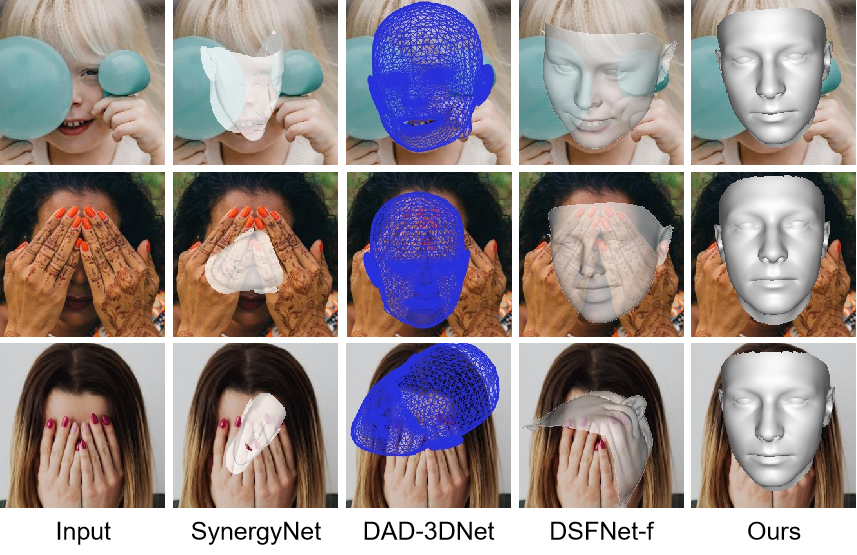}
	\vspace{-25pt}
	\caption{Qualitative comparison results of front-range head pose estimation (HPE) methods. Our method is more robust to severe occlusion. Images are taken from the paper DSFNet-f \cite{li2023dsfnet}.}
	\label{Comparing1}
	\vspace{-10pt}
\end{figure}

\begin{figure}[t]
	\centering
	\includegraphics[width=\columnwidth]{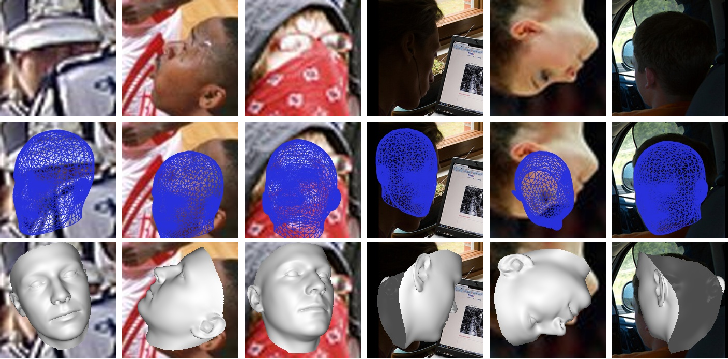}
	\vspace{-20pt}
	\caption{Qualitative comparison results of full-range head pose estimation (HPE) methods on wild head images. Our method is more robust to severe occlusion, atypical pose and invisible face than DAD-3DNet \cite{martyniuk2022dad}.}
	\label{Comparing2}
	\vspace{-10pt}
\end{figure}

\begin{figure*}[t]
	\centering
	\includegraphics[width=\textwidth]{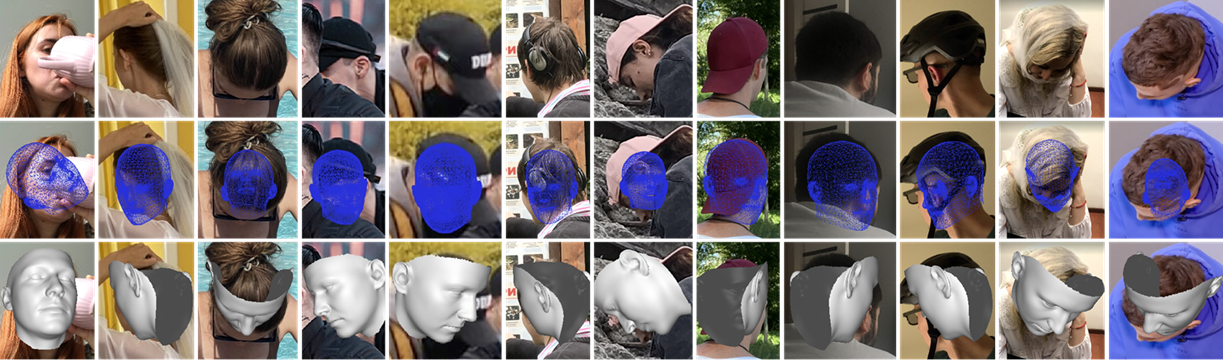}
	\vspace{-20pt}
	\caption{Qualitative comparison results of full-range head pose estimation (HPE) between our method (\textit{3rd line}) and DAD-3DNet \cite{martyniuk2022dad} (\textit{2nd line}). All head images are from DAD-3DHeads test-set (\textit{1st line}), which never appeared during SSL training.}
	\label{ComparingDAD}
	\vspace{-10pt}
\end{figure*}

\begin{figure}[t]
	\centering
	\includegraphics[width=\columnwidth]{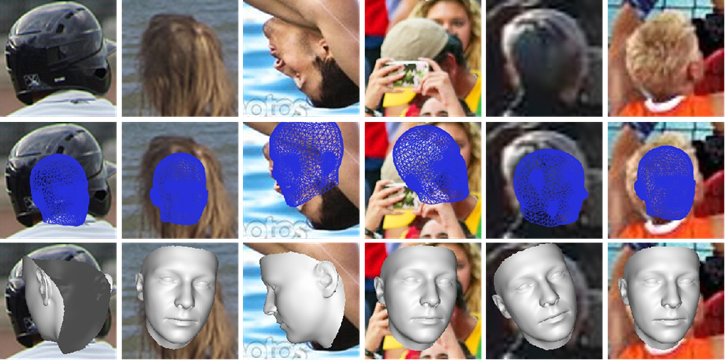}
	\vspace{-20pt}
	\caption{Failure cases (\textit{third line}) of our SemiUHPE on some wild challenging heads (\textit{fisrt line}). These samples are also very hard for DAD-3DNet \cite{martyniuk2022dad} (\textit{second line}) to deal with.}
	\label{FailureCases}
	\vspace{-10pt}
\end{figure}

\subsection{Qualitative Comparison}

To further explain the superiority of our SemiUHPE, we present the qualitative comparison with SOTA counterparts including SynergyNet \cite{wu2021synergy}, DAD-3DNet \cite{martyniuk2022dad} and DSFNet-f \cite{li2023dsfnet}. As shown in Fig.~\ref{Comparing1} and Fig.~\ref{Comparing2}, our method can obtain impressive HPE results on front yet severely occluded faces and in-the-wild challenging heads. Moreover, we present comparison on never-before-experienced yet challenging samples from the test-set of DAD-3DHeads. As shown in Fig.~\ref{ComparingDAD}, although these images have high-definition, DAD-3DNet may make significant mistakes, leading to disordered results of head reconstruction. While, our method usually gives a satisfactory estimation. More convincing qualitative results can be found in our project link.

Although the overall effect is impressive, our SemiUHPE may still fail on some cases of heavy blur, invisible face, severe occlusions, or atypical pose (like faces upside-down). Sometimes, there is more than one challenge, as shown in Fig.~\ref{FailureCases} of the third case (atypical pose + self-occlusion) and the last two cases (blurry + backward). Although we humans have strong prior knowledge to help infer the poses of challenging heads, the network cannot do this so far. We believe that in these situations, it is necessary for the model to rely on the context of the human body to identify its head pose, especially for the first, second and fourth cases in Fig.~\ref{FailureCases}, where heads facing back and completely covered by other objects such as hats and hair.

\subsection{Ablation Studies}
In this part, we give detailed studies for explaining the effect of our proposed three strategies. Then, we present the studies of dynamic thresholds changing and unsupervised convergence tendency, respectively.

\setlength{\tabcolsep}{1.6pt}
\begin{table}[t]\scriptsize  
	\begin{center}
	\caption{Euler angles errors on AFLW2000. Models are trained on 300W-LP with different input cropping ways and pose rotation representations.}
	\vspace{-10pt}
	\begin{tabular}{l|c|c|c|cccc}
	\Xhline{1.2pt}
	Method & Backbone & Cropping & Rot-Rep & Pitch & Yaw & Roll & MAE \\
	\Xhline{1.2pt}
	FSA-Net \cite{yang2019fsa} & ResNet50 & Naive & Euler angles & 6.08 & 4.50 & 4.64 & 5.07 \\
	FSA-Net$\dag$  \cite{yang2019fsa} & ResNet50 & Ours & Euler angles & 5.42 & 4.01 & 3.75 & 4.39 \\
	6DRepNet \cite{hempel20226d} & RepVGG & Naive & trivial matrix & 4.91 & 3.63 & 3.37 & 3.97 \\
	6DRepNet$\dag$ \cite{hempel20226d} & RepVGG & Ours & trivial matrix & 4.58 & 3.04 & 2.86 & 3.49 \\
	Supervised & ResNet50 & Ours & matrix Fisher & 4.58 & 3.20 & 2.95 & 3.58 \\
	Supervised & RepVGG & Ours & matrix Fisher & 4.50 & 3.18 & 2.81 & 3.50 \\
	Supervised & EffNetV2-S & Ours & matrix Fisher & \textcolor{red}{4.40} & \textcolor{red}{2.99} & \textcolor{red}{2.75} & \textcolor{red}{3.38} \\
	\Xhline{1.2pt}
	\end{tabular}
	\label{tabFour}
	\end{center}
	\vspace{-10pt}
\end{table}

\begin{figure*}[t]
	\centering
	\subfloat[]{\includegraphics[width=0.33\textwidth]{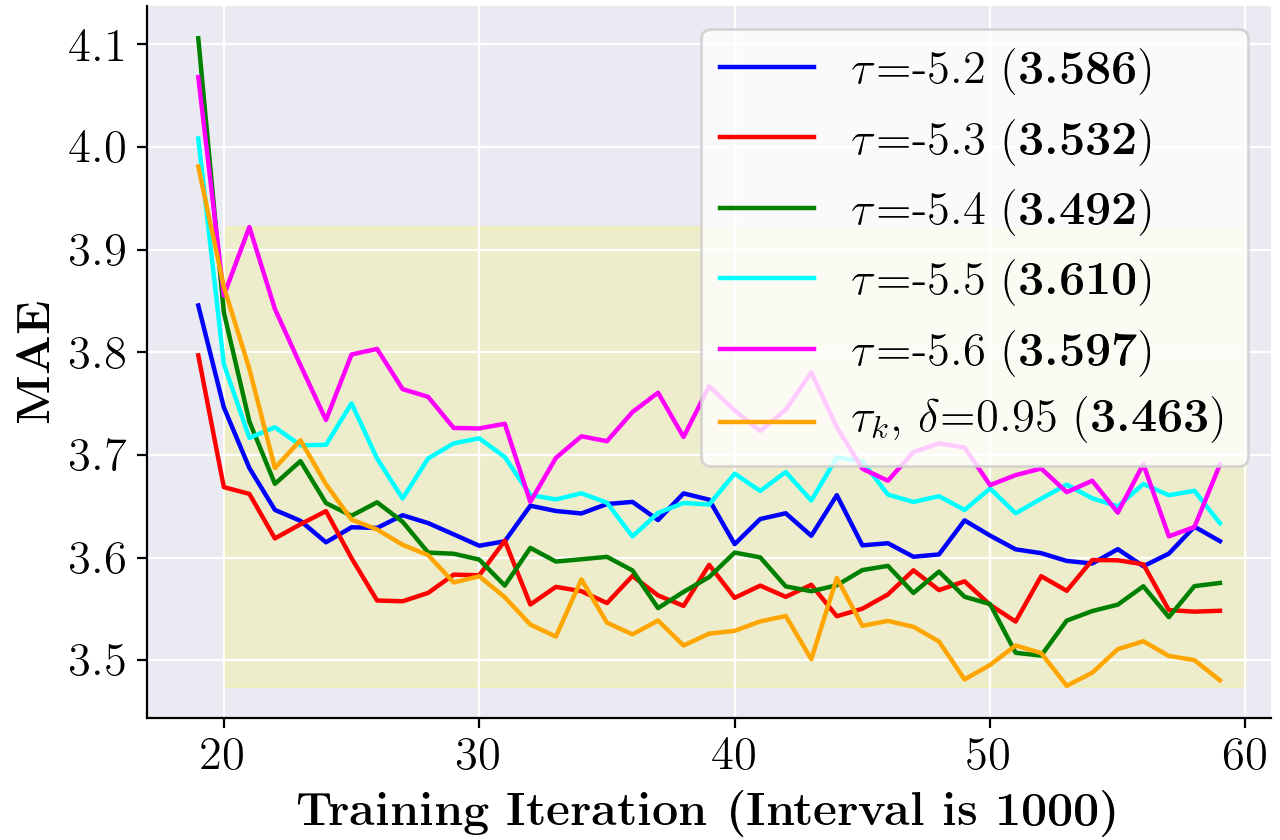}
	\label{asExpsA}}
	\subfloat[]{\includegraphics[width=0.33\textwidth]{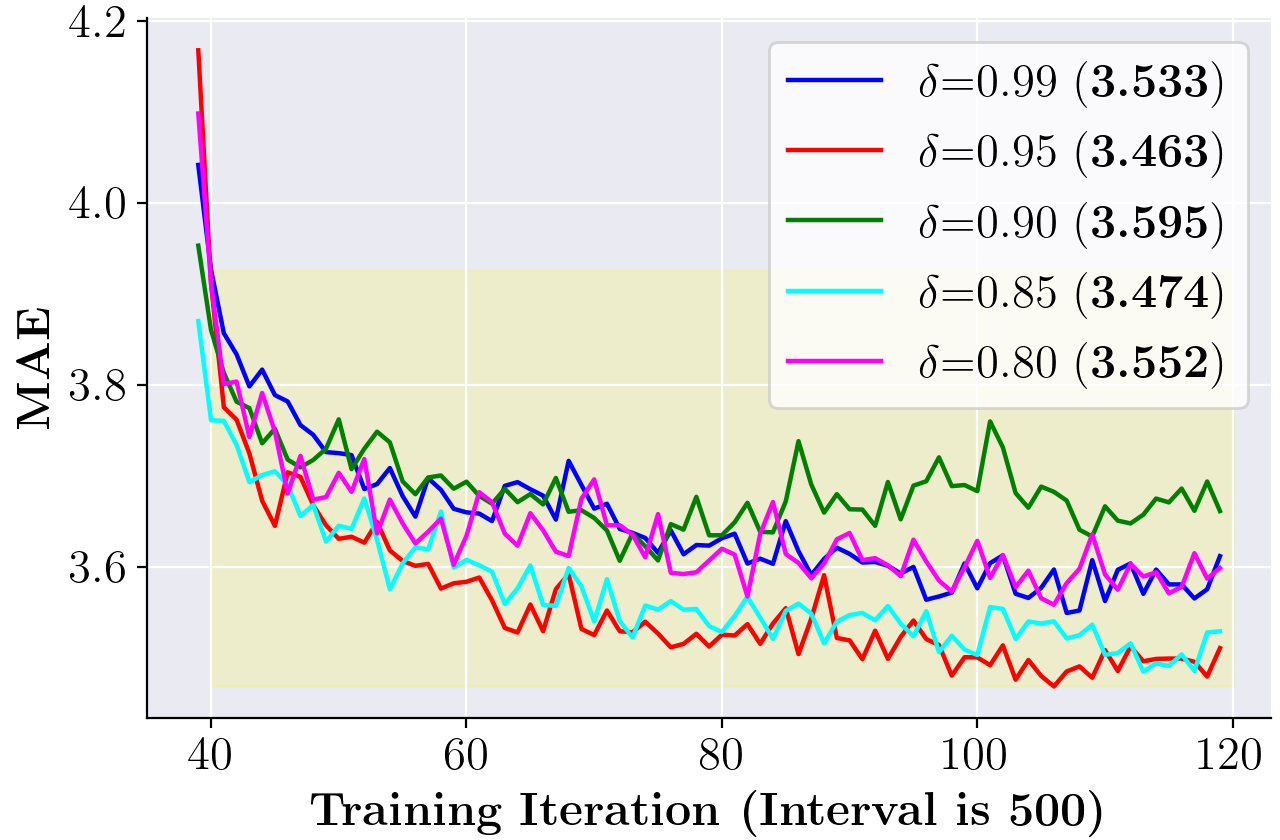}
	\label{asExpsB}}
	\subfloat[]{\includegraphics[width=0.33\textwidth]{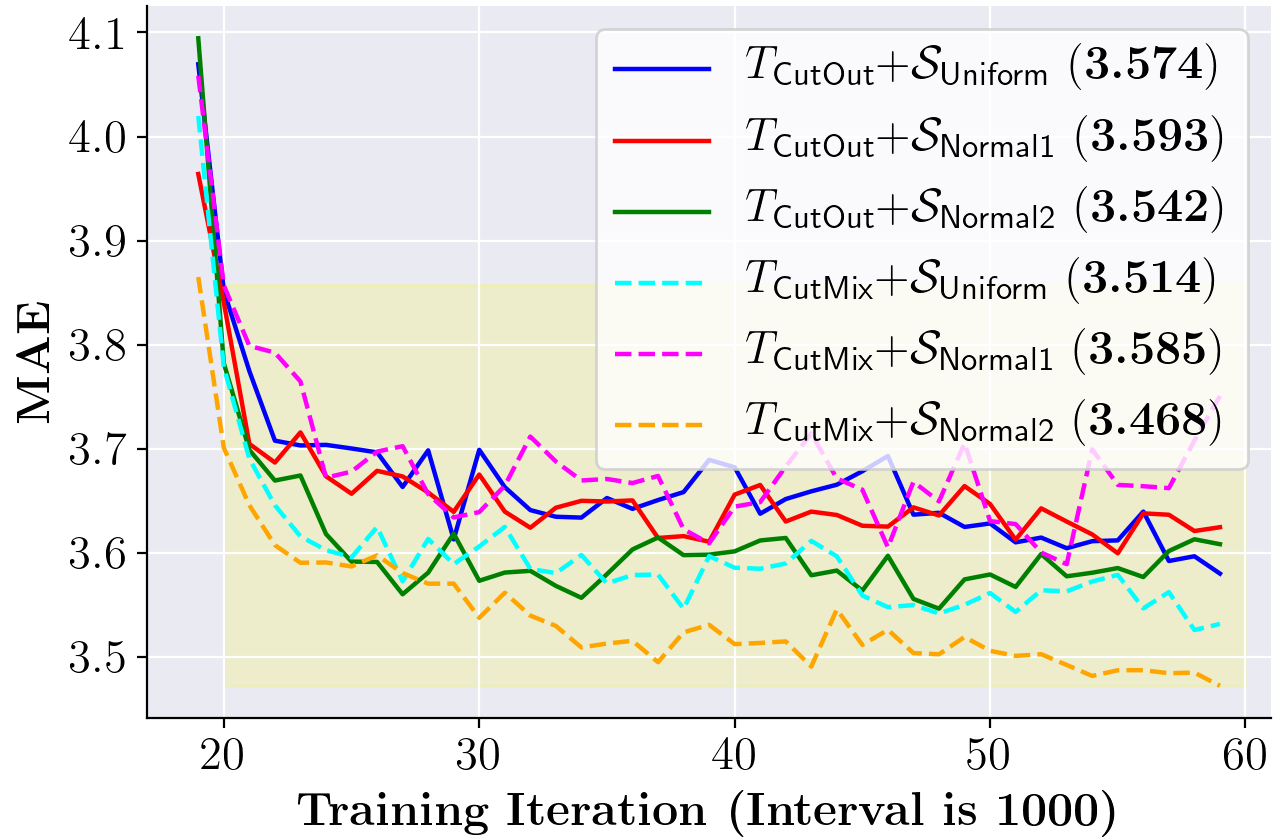}
	\label{asExpsC}}\\
	\vspace{-10pt}
	\subfloat[]{\includegraphics[width=0.33\textwidth]{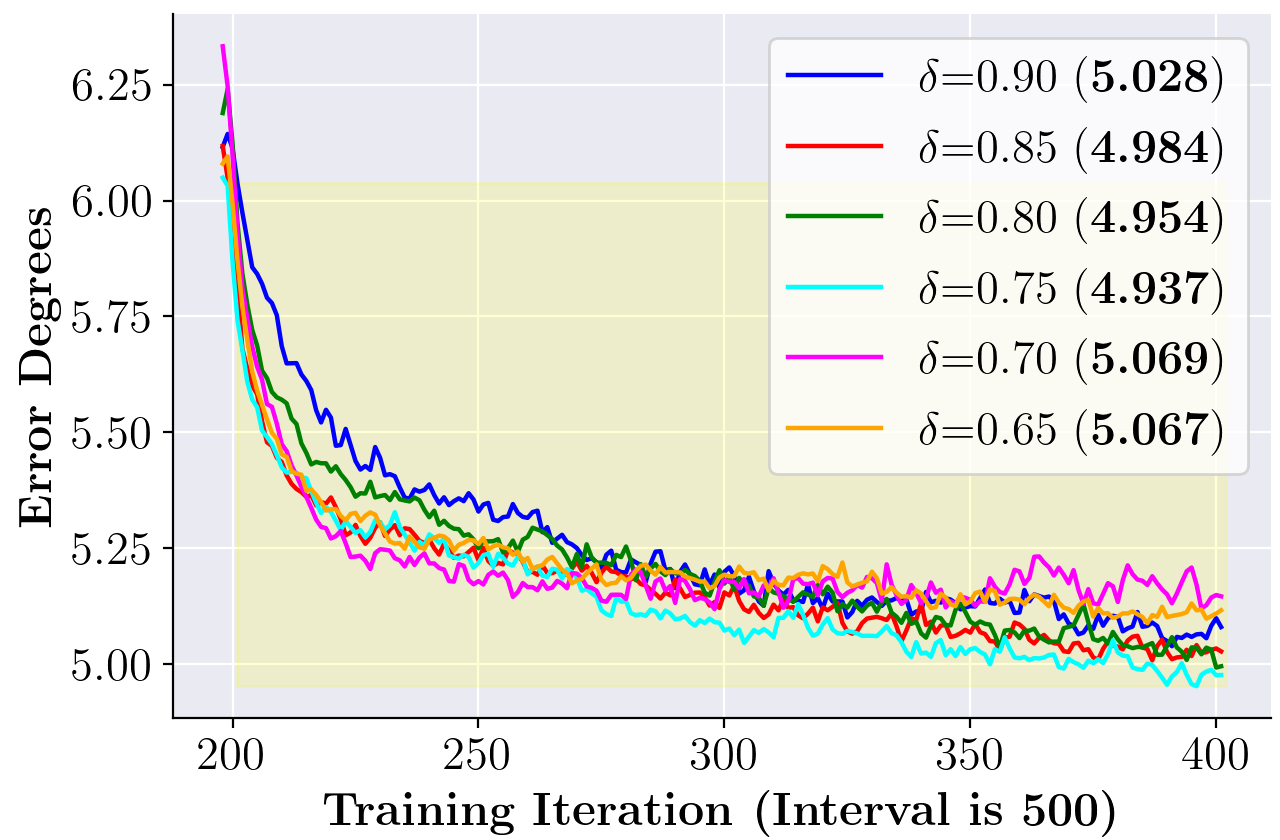}
	\label{asExpsD}}
	\subfloat[]{\includegraphics[width=0.33\textwidth]{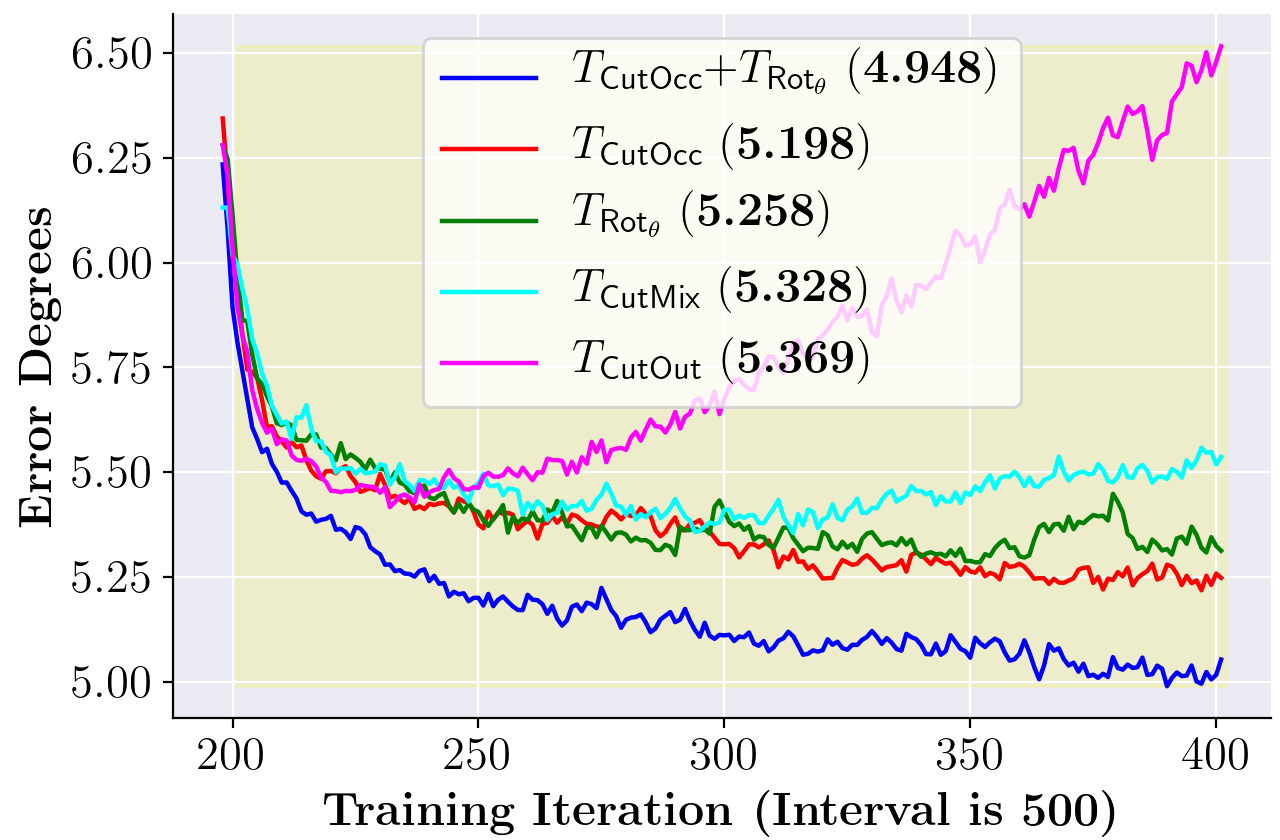}
	\label{asExpsE}}
	\subfloat[]{\includegraphics[width=0.33\textwidth]{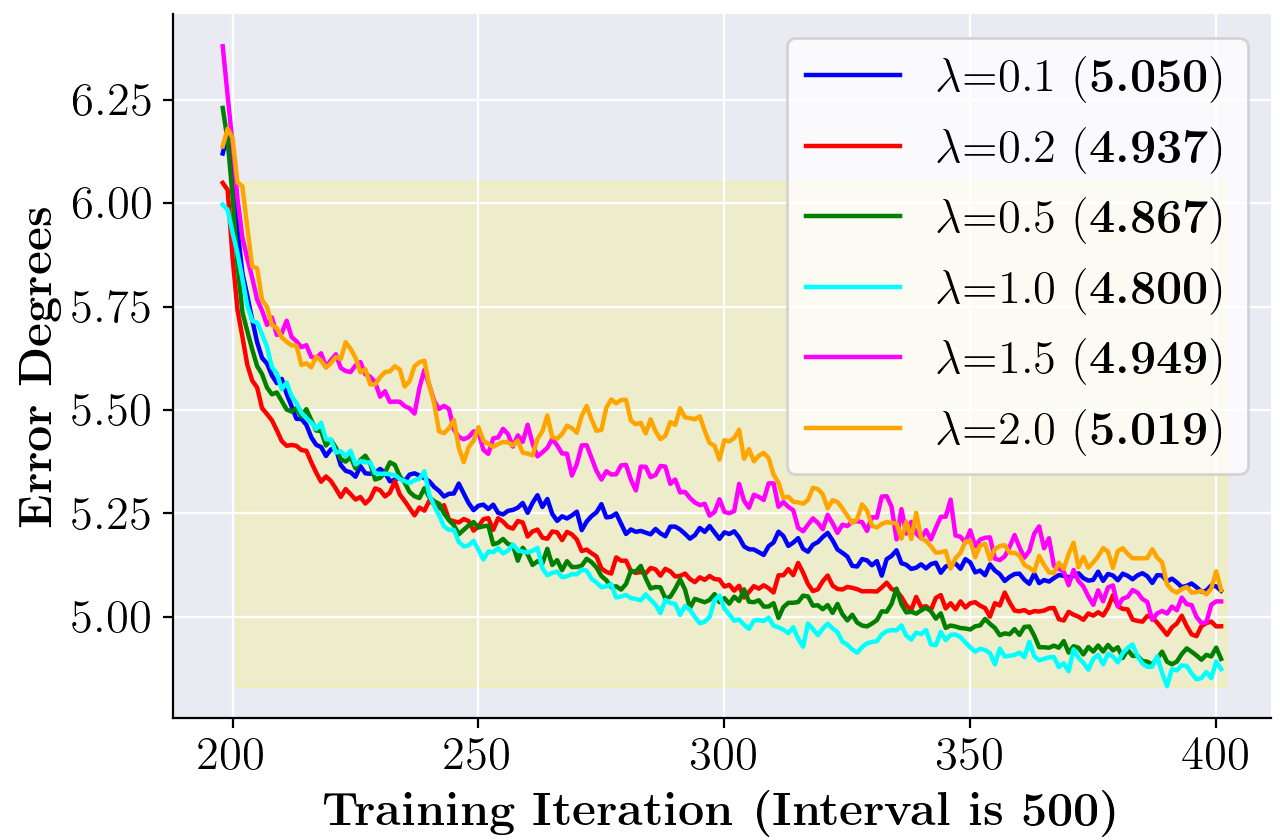}
	\label{asExpsF}}
	\vspace{-10pt}
	\caption{The backbone of each model is ResNet50. The bold number within parentheses is the best result. We mainly show results in \textsf{Phase2}. \textbf{Top Row:} \textbf{Setting1} with 20\% labels. (a) The comparison of using a dynamic threshold $\tau_k$ with $\delta\!=\!0.95$ or pre-fixed threshold $\tau$. (b) The influence of $\delta$. (c) The effect of different sampling ways. \textbf{Bottom Row:} \textbf{Setting3}. (d) The influence of $\delta$ with $\lambda\!=\!0.2$. (e) The effect of different augmentations with $\lambda\!=\!0.2$ and $\delta\!=\!0.75$. (f) The influence of $\lambda$ with $\delta\!=\!0.75$.}
	\vspace{-15pt}
	\label{asExps}
\end{figure*}

\begin{figure}[t]
	\centering
	\subfloat[]{\includegraphics[width=0.495\columnwidth]{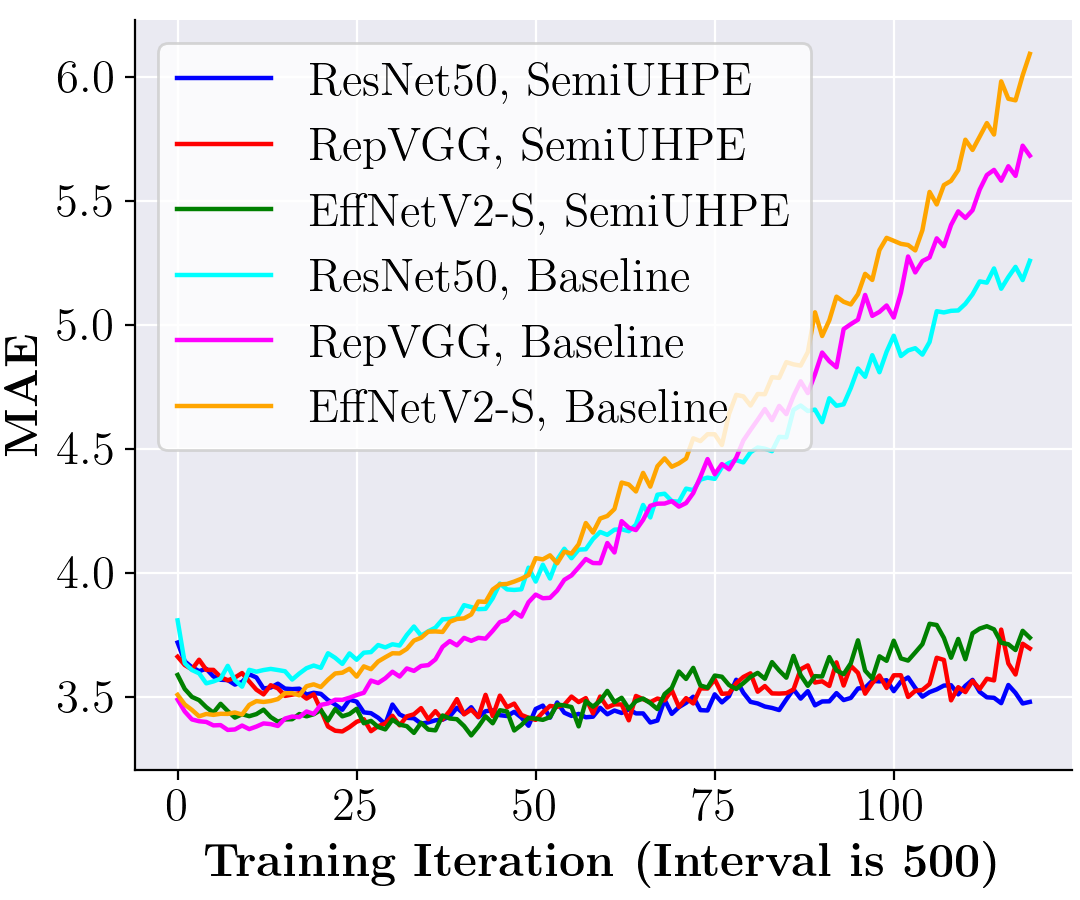}
	\label{asExpsAdd2A}}
	\subfloat[]{\includegraphics[width=0.495\columnwidth]{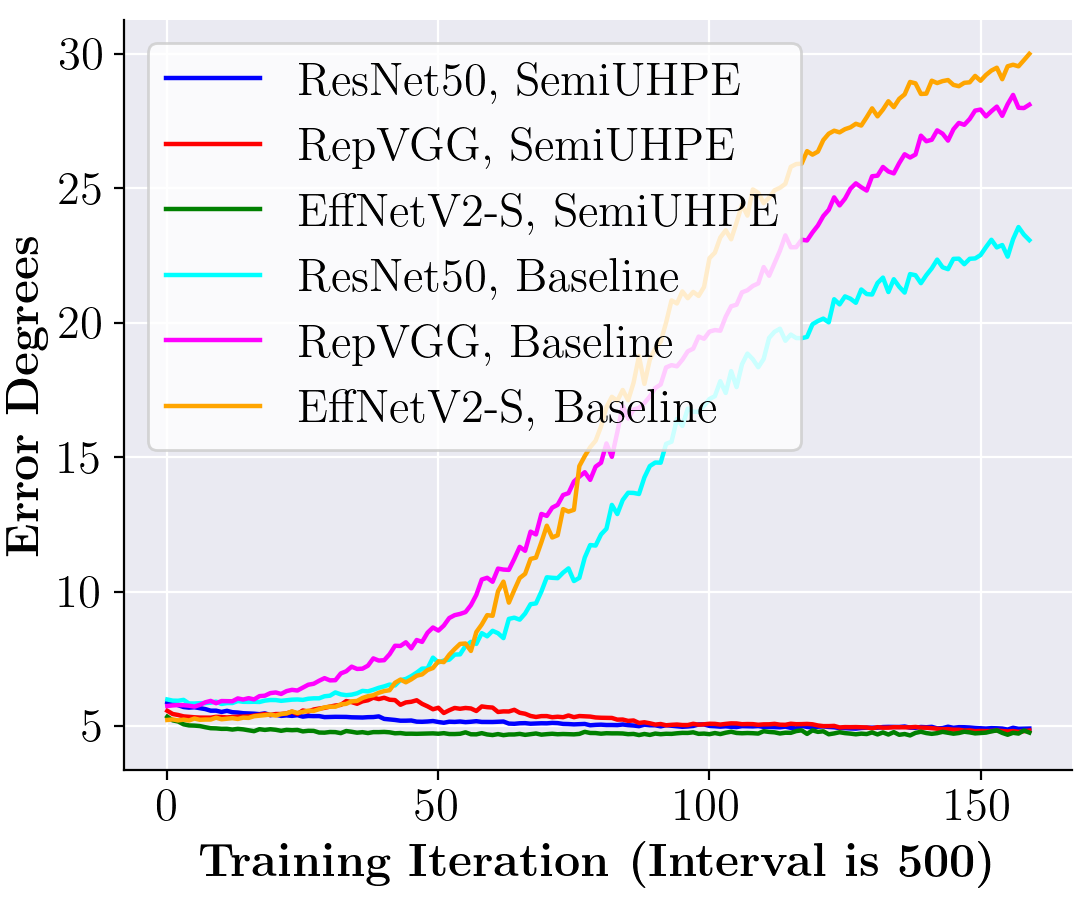}
	\label{asExpsAdd2B}}
	\vspace{-10pt}
	\caption{Testing results of the baseline FisherMatch and our proposed SemiUHPE in \textbf{Setting2} (a) and \textbf{Setting3} (b). Only training steps in \textsf{Phase2} are shown for clearer comparison.}
	\label{asExpsAdd2}
	\vspace{-5pt}
\end{figure}

\textbf{Aspect-Ratio Invariant Cropping.}\label{as01}
We took FSA-Net \cite{yang2019fsa} and 6DRepNet \cite{hempel20226d} using the naive cropping-resizing way for comparing. Then, we replaced them with aspect-ratio invariant cropping, and retrained new versions FSA-Net$\dag$ and 6DRepNet$\dag$. We also listed results of the trivial supervised method. As shown in Table~\ref{tabFour}, both the original FSA-Net and 6DRepNet are significantly improved. When using RepVGG and new cropping, 6DRepNet$\dag$ can even compete with the supervised method equipped an advanced matrix Fisher representation. These prove the superiority of this simple yet efficient size-invariant preprocessing.

\textbf{Dynamic Entropy-based Filtering.}\label{as02}
We designed three groups of experiments for explaining the effectiveness of dynamic filtering strategy. Firstly, in Fig.~\ref{asExpsA}, we searched for the optimal pre-fixed threshold $\tau$ used by FisherMatch. Nonetheless, our dynamic threshold $\tau_k$ with $\delta\!=\!0.95$ got the best result. We also searched for the optimal $\delta$ for different unlabeled datasets. Then, in Fig.~\ref{asExpsB}, we got the optimal $\delta\!=\!0.95$, which is a large ratio due to that both labeled and unlabeled data are in 300W-LP. Finally, in Fig.~\ref{asExpsD}, we got a lower optimal $\delta\!=\!0.75$, which is caused by more difficult and noisy heads in the unlabeled COCOHead. Usually, an ideal $\delta$ should equal to $1\!-\!\|\mathcal{D}^u_{ood}\|/\|\mathcal{D}^u\|$. But we cannot obtain the accurate ratio of $\mathcal{D}^u_{ood}$ in $\mathcal{D}^u$. We thus estimated a suitable $\delta$ by ablation studies. For example, the optimal $\delta$ of COCOHead is 0.75 in \textbf{Setting2} and \textbf{Setting3}, which means $\delta$ may be dataset-related yet not sensitive to task settings.

\textbf{Head-Oriented Strong Augmentations.}\label{as03}
We conducted two groups of experiments for showing the effectiveness of new strong augmentations. In Fig.~\ref{asExpsC}, we can see that CutMix is always better than CutOut. The best sampling distribution is $\mathcal{S}_\mathsf{Normal2}$ for its reasonable concentration of occlusion generation. In Fig.~\ref{asExpsE}, we kept using $\mathcal{S}_\mathsf{Normal2}$ and observed the same effect of CutOut and CutMix. When combining them together, the new $T_\mathsf{CutOcc}$ can further reduce HPE errors. Independently, the proposed rotation consistency augmentation $T_{\mathsf{Rot}_\theta}$ can also improve performance. Finally, when applying both $T_\mathsf{CutOcc}$ and $T_{\mathsf{Rot}_\theta}$, we achieved the best result with a remarkable promotion. 

\textbf{Influence of Unsupervised Loss Weight.}\label{as04}
As shown in Fig.~\ref{asExpsF}, we selected the unsupervised loss weight $\lambda$ from the list (0.1, 0.2, 0.5, 1.0, 1.5, 2.0). Our method performed best when setting $\lambda\!=\!1.0$, which indicates that it does not require careful adjustments of the unsupervised part weight and has a robust performance about this hyper-parameter.


\textbf{Convergence Curve of Baseline FisherMatch.}\label{as06}
As shown in Fig.~\ref{asExpsAdd2}, we plotted convergence curves of the baseline FisherMatch as well as our proposed SemiUHPE. It is obvious that whether it is in \textbf{Setting2} (see Fig.~\ref{asExpsAdd2A}) or \textbf{Setting3} (see Fig.~\ref{asExpsAdd2B}), the baseline method always begins to collapse after quickly reaching an optimal point with using either backbone. We assume that this is caused by the domain differences between labeled and unlabeled datasets. Our proposed SemiUHPE can significantly alleviate this problem and converge relatively smoothly.

\subsection{Other Optional Setups}

\setlength{\tabcolsep}{1pt}
\begin{table}[!t]  
	\begin{center}
	\caption{The HPE results on the \textit{val-set} of dataset DAD-3DHeads by using different unlabeled datasets.}
	\vspace{-10pt}
	\begin{tabular}{l|c|c|c}
	\Xhline{1.2pt}
	Method & \makecell{Unlabeled \\ Dataset} & {$\|\mathbf{I}\!-\!\mathbf{R}_1\mathbf{R}_2^T\|_F\downarrow$} & \makecell{Angle error\\ (degree)$\downarrow$} \\
	\Xhline{1.2pt}
	DAD-3DNet \cite{martyniuk2022dad} & --- & 0.130 & 5.456 \\
	\hline
	\multirow{4}{*}{\makecell{SemiUHPE\\(ours)}} & WiderFace & 0.127 & 5.300$_{(\;\;2.9\%\downarrow)}$ \\
	~ & CrowdHuman & 0.125 &  5.226$_{(\;\;4.2\%\downarrow)}$ \\
	~ & OpenImageV6 & 0.123 &  5.174$_{(\;\;5.2\%\downarrow)}$ \\
	~ & COCOHead & 0.116 & 4.800$_{(12.0\%\downarrow)}$ \\
	\Xhline{1.2pt}
	\end{tabular}
	\label{OptionalTabA}
	\end{center}
	\vspace{-15pt}
\end{table}

\setlength{\tabcolsep}{3pt}
\begin{table}[!t]  
	\begin{center}
	\caption{Euler angles errors on AFLW2000. Models are trained on 300W-LP with different ratios of label.}
	\vspace{-10pt}
	\begin{tabular}{c|l|c|ccccc}
	\Xhline{1.2pt}
	Type & Method & Backbone & 2\% & 5\% & 10\% & 20\% & All \\
	\Xhline{1.2pt}
	\multirow{2}{*}{SL} & Sup. & TinyViT-22M & 4.116 & 3.734 & 3.677 & 3.512 & 3.455 \\
	~ & Sup. & EffNetV2-S & 4.009 & 3.678 & 3.517 & 3.444 & 3.379 \\
	\hline
	\multirow{4}{*}{SSL} & \tikzmark{bbv}Base. & TinyViT-22M & 4.031 & 3.691 & 3.464 & 3.415 & --- \\
	~ & \tikzmark{ddv}Base. & EffNetV2-S & 3.991 & 3.596 & 3.448 & 3.372 & --- \\
	~ & \tikzmark{aav}Ours & TinyViT-22M & 3.914 & 3.647 & 3.514 & 3.434 & --- \\
	~ & \tikzmark{ccv}Ours & EffNetV2-S & 3.835 & 3.526 & 3.377 & 3.348 & --- \\
	\Xhline{1.2pt}
	\end{tabular}
    	\begin{tikzpicture}[ remember picture, overlay, thick]
   	 	\draw [<-,green] ([xshift=-0.5ex, yshift=1ex]pic cs:bbv) [bend right] to ([xshift=-0.5ex, yshift=1ex]pic cs:aav);
    		\draw [<-,magenta] ([xshift=-0.5ex, yshift=1ex]pic cs:ddv) [bend right] to ([xshift=-0.5ex, yshift=1ex]pic cs:ccv);
	\end{tikzpicture}
	\label{OptionalTabB}
	\end{center}
	\vspace{-10pt}
\end{table}

\subsubsection{Choosing Different Unlabeled Datasets.}
As discussed in the previous content, in addition to COCOHead \cite{lin2014microsoft}, there are also other alternative unlabeled datasets such as WiderFace \cite{yang2016wider}, CrowdHuman \cite{shao2018crowdhuman} and OpenImageV6 \cite{xie2020unsupervised} that contain many wild heads. A natural doubt is, will applying these similar substances yield better results than using COCOHead. To answer this question, we followed the steps of generating COCOHead, processed original WiderFace and CrowdHuman datasets, and obtained the corresponding unlabeled head sets. Specifically, we removed samples with head size smaller than 25, 30 and 30 pixels in WiderFace, CrowdHuman and OpenImageV6. Then, we got about 62K, 163K and 165K heads, respectively. Then, we followed \textbf{Setting3} to implement the similar \textbf{DAD-WiderFace}, \textbf{DAD-CrowdHuman} and and \textbf{DAD-OpenImage} experiments. Considering that both CrowdHuman and OpenImageV6 have about $2\times$ samples than COCOHead ($\sim$74K) or WiderFace, we adjusted their iterations in \textsf{Phase2} from 100K into 200K. Without loss of generality, all compared models used the ResNet50 \cite{he2016deep} as their network backbones. 

As shown in Table \ref{OptionalTabA}, although our method SemiUHPE using either unlabeled dataset can surpass the supervised DAD-3DNet \cite{martyniuk2022dad}, it has significant discrepancies when applying different unlabeled datasets. For the angle error, our method can improve DAD-3DNet by 12.0\% when using COCOHead, which is much more prominent than using WiderFace (2.9\%), CrowdHuman (4.2\%) or OpenImageV6 (5.2\%). This is understandable because most human heads in these datasets are face-visible, especially the WiderFace originally built for face detection task. Besides, CrowdHuman and OpenImageV6 have many harmful head bounding box annotations which are illegal or unrecognizable. In summary, COCOHead is the most suitable unlabeled choice. In practical applications, if we want to pursue higher performance, we can also choose to merge these datasets for co-training (such as the last row in Table~\ref{tabThree}).

\subsubsection{Choosing Transformer-based Backbones}
Although we have chosen three different backbones including ResNet50 \cite{he2016deep}, RepVGG \cite{ding2021repvgg} and EfficientNetV2-S \cite{tan2021efficientnetv2} to conduct extensive experiments, these networks are all based on CNNs, which have been challenged in recent years by transformer-based alternatives \cite{dosovitskiy2020image, liu2021swin, liu2022swin}. Actually, the fully supervised method TokenHPE \cite{zhang2023tokenhpe, liu2023orientation} has initially revealed the great potential of using transformer networks (\eg, the ViT-Base/16 \cite{dosovitskiy2020image}) to deal with the HPE task. Therefore, we also considered whether we could use a more superior transformer-based backbone to boost SemiUHPE. After considering various aspects, we decided to adopt TinyViT-22M \cite{wu2022tinyvit} as a trial. On the one hand, TinyViT-22M has similar parameters to our used CNN-based networks, and on the other hand, its pre-trained model has higher classification accuracy yet smaller calculation amount than the original ViT-Base/16. This means it is more likely to achieve better HPE results.

Specifically, with using TinyViT-22M as the backbone, we followed the \textbf{Setting1} and implemented all three methods including Sup., Base and SemiUHPE. As shown in Table~\ref{OptionalTabB}, although TinyViT-22M has a similar top-1 accuracy with EffNetV2-S on ImageNet (84.8\% vs. 84.9\%), its performance can be much worse than EffNetV2-S when trained with full or semi-supervision. As a reasonable reference, in Table~\ref{tabTwo}, the transformer-based TokenHPE \cite{zhang2023tokenhpe} also failed to achieve lower HPE errors than many CNN-based counterparts. And our reproduced supervised method using the TinyViT-22M backbone and matrix Fisher representation can get quite low results (MAE=3.46) if comparing with methods in Table~\ref{tabTwo}. In particular, when using TinyViT-22M for SSL training, our method performed worse than the baseline method when the labeling rate is 10\% or 20\%. This anomaly is very different from the case when using other backbones. We suspect that this is because the transformer structure needs to preprocess the input image into smaller patches, which conflicts with the operations such as CutOut \cite{devries2017improved} and CutMix \cite{yun2019cutmix} used in the strong augmentations on unlabeled images. Nevertheless, we envision that these troubles may be alleviated by applying data augmentation strategies specifically designed for vision transformer families\cite{chen2022transmix, liu2022tokenmix, fang2024feataug}, which can be considered as a trivial extension of our work. On the other hand, it has been widely proven that vision transformers require a large amount of labeled data to train in order to unleash their scaling laws, such as for 3D face reconstruction \cite{zhang2023accurate} and 3D hand reconstruction \cite{pavlakos2024reconstructing}. But as we have revealed, this does not seem to benefit HPE tasks in SSL settings.

\section{Conclusion and Discussion}\label{conclusion}

In this paper, we aim to address the unconstrained head pose estimation task on less-touched wild head images. Due to the lack of corresponding labels, we turn to semi-supervised learning techniques. Based on empirically effective frameworks, we propose the dynamic entropy-based filtering for gradually updating thresholds and head-oriented strong augmentations for better enforcing consistency training. By combining the proposed aspect-ratio invariant cropping, our method can achieve optimal HPE performance quantitatively and qualitatively on omnidirectional wild heads. We also demonstrate the scalability and versatility of SemiUHPE on generic object rotation regression and 3D head reconstruction. We expect that our work will greatly inspire related downstream applications. 

Last but not least, although our method has achieved stunning results in estimating the pose of wild head, there are still many aspects worth exploring in depth. We summarize the possible perspectives for future research as follows.

\begin{itemize}
\item{\textbf{Domain gaps among datasets.} Strictly speaking, there are conspicuous differences between labeled and unlabeled datasets we used in this paper. For example, the labeled 300W-LP is artificially synthesized through the face profiling algorithm with artifacts, while the unlabeled COCOHead is collected in the wild containing realistic and natural samples. We may utilize domain adaptation strategies to alleviate this problem.}

\item{\textbf{Combination with vision-language models.} Recently, large vision-language models (VLMs) and multimodal LLMs (mLLMs) have demonstrated strong generalizable visual reasoning abilities by aligning image and text inputs. Beyond using text descriptions to mitigate data scarcity \cite{wang2024language} or to improve robustness in HPE \cite{tian2024hpe}, a more concrete strategy is to cast unconstrained HPE into a visual question answering (VQA) task. For example, both the cropped head image and its reconstructed 3D mesh can be provided to an mLLM, together with a query about the plausibility of the estimated pose. If the mLLM detects a severe error, the system can re-estimate after small perturbations or reject the prediction. Our preliminary test with Gemini 2.5 Flash shows that it can reliably flag most obvious errors, suggesting a promising future direction.}


\item{\textbf{More accurate head pose estimation.} Our method still cannot address some hard cases shown in Fig.~\ref{FailureCases}, where most facial features are missing due to severe occlusion or backward orientation. A potential remedy is to leverage contextual cues beyond the cropped head. For instance, the orientation of the upper body \cite{nonaka2022dynamic, zhou2022joint} can provide a strong prior: if the torso clearly faces away from the camera, the head pose should be consistent with a back-of-head orientation. Similarly, surrounding information such as hair-dominated regions with absent facial features \cite{nakatani2023interaction, zhou2023directmhp} can help detect backward heads where the yaw absolute angle must exceed $90^{\circ}$. Integrating such priors may effectively reduce catastrophic errors that cannot be solved by single-frame appearance cues alone.}

\item{\textbf{Video-level HPE extension.} While our SemiUHPE current focuses on single-frame, applying it to video sequences may lead to temporal jitter or abrupt errors, especially for backward-facing heads. A promising direction is to incorporate temporal consistency constraints, ensuring smooth pose transitions across consecutive frames. Another possibility is to exploit short-term temporal windows, where neighboring frames provide complementary cues to disambiguate extreme poses. Such extensions could further enhance the robustness of SemiUHPE in real-world video applications.}

\item{\textbf{Video-based SemiUHPE.} Another promising extension is to move from single-frame SemiUHPE to video-based semi-supervised head pose estimation. By redesigning the backbone to process short frame sequences, the model could implicitly capture temporal head-motion cues and output smooth 3D pose trajectories. A teacher–student framework built on annotated video datasets (e.g., BIWI \cite{fanelli2013random}, CMU Panoptic \cite{joo2015panoptic}) together with large-scale unlabeled online videos would provide a natural baseline. This direction also parallels recent advances in sequential visuomotor  imitation learning, where temporal visual observations are mapped to continuous 6DoF trajectories, suggesting that frameworks such as ACT \cite{zhao2023learning} or Diffusion Policy \cite{chi2023diffusion} could be adapted for head pose estimation.}

\item{\textbf{Impact on downstream tasks.} Our SemiUHPE also has implications for broader head-related tasks. For 3D head generation, it can help diagnose and re-balance pose distributions in training datasets, mitigating bias caused by underrepresented large poses. For talking-head generation, robust estimation of extreme angles may reduce jittering artifacts in synthesized videos. For face-swapping, accurate pose priors can guide the alignment of wrapped facial textures, alleviating failures under large yaw angles. These extensions highlight the potential of SemiUHPE as a general building block for robust head-driven applications.}


\end{itemize}




\bibliographystyle{IEEEtran}
\bibliography{refs}

\end{document}